%% file: arxiv.tex
\pdfoutput=1
\documentclass{article}

\usepackage{microtype}
\usepackage{graphicx}
\usepackage{subcaption}
\usepackage{booktabs} %

\usepackage{hyperref}

\usepackage[table]{xcolor}

\usepackage[accepted]{icml2025}
\usepackage{listings}

\usepackage{amsmath}
\usepackage{amssymb}
\usepackage{mathtools}
\usepackage{amsthm}
\usepackage{framed}
\usepackage{xspace}
\usepackage{booktabs}
\usepackage{multirow}
\usepackage{tabularx}
\usepackage{array}
\usepackage{enumitem}
\usepackage{arydshln}

\usepackage{tikz}
\usepackage{pgfplots}
\usetikzlibrary{shapes.geometric, arrows, calc}
\usetikzlibrary{arrows.meta, positioning}
\usetikzlibrary{decorations.text}
\usetikzlibrary{shapes.arrows}
\usetikzlibrary{patterns}

\newlength{\figurewidth}
\newlength{\figureheight}

\usepackage[capitalize,nameinlink]{cleveref}
\crefname{section}{Sec.}{Secs.}
\crefname{table}{Table}{Tables}
\crefname{figure}{Fig.}{Figs.}
\crefname{appendix}{App.}{Apps.}
\crefname{algorithm}{Alg.}{Algs.}

\usepackage{colortbl}
\usepackage{xcolor}
\usepackage{pifont}
\usepackage{array}
\usepackage{wrapfig}

\definecolor{mydarkblue}{rgb}{0,0.08,0.45} 
\definecolor{primarycolor}{HTML}{0000cc} 
\colorlet{highlight}{primarycolor!05!white} 

\theoremstyle{plain}

\theoremstyle{definition}

\theoremstyle{remark}

\usepackage{xspace}
\newcommand{\ours}{Plan$^\ast$RAG\xspace}%
\newcommand{\planrag}{Plan$^\ast$RAG\xspace}

\usepackage[textsize=tiny]{todonotes}

\newcommand{\ie}{\textit{i.e.}\xspace}

\newcommand{\cf}{\textit{cf.}\xspace}

\newcommand{\fllm}{\mathbf{f}_\textrm{\tiny LLM}}

\newcommand{\cmark}{\textcolor{green!50!black}{\ding{51}}}%
\newcommand{\xmark}{\textcolor{black!30!red}{\ding{55}}}%
\newcolumntype{H}{>{\columncolor{highlight}}c}
\renewcommand{\paragraph}[1]{\noindent\textbf{#1}~~}

\icmltitlerunning{Efficient Test-Time Planning for Retrieval Augmented Generation}

\begin{document}

\twocolumn[
\icmltitle{Plan$^\ast$RAG: Efficient Test-Time Planning for Retrieval Augmented Generation}

\icmlsetsymbol{equal}{$\dagger$}

\begin{icmlauthorlist}
\icmlauthor{Prakhar Verma}{equal,aalto}
\icmlauthor{Sukruta Prakash Midigeshi}{msr}
\icmlauthor{Gaurav Sinha}{msr}
\icmlauthor{Arno Solin}{aalto}\\
\icmlauthor{Nagarajan Natarajan}{msr}
\icmlauthor{Amit Sharma}{msr}
\end{icmlauthorlist}

\icmlaffiliation{aalto}{Aalto University, Finland}
\icmlaffiliation{msr}{Microsoft Research}

\icmlcorrespondingauthor{}{prakhar.verma@aalto.fi}
\icmlcorrespondingauthor{}{amshar@microsoft.com}

\icmlkeywords{RAG, LLM, Reasoning, Planning}

\vskip 0.3in
]

\printAffiliationsAndNotice{\icmlEqualContribution} %
\begin{abstract}
We introduce \ours, a novel framework that enables structured multi-hop reasoning in retrieval-augmented generation (RAG) through test-time reasoning plan generation. While existing approaches such as ReAct maintain reasoning chains within the language model's context window, we observe that this often leads to plan fragmentation and execution failures. Our key insight is that by isolating the reasoning plan as a directed acyclic graph (DAG) outside the LM's working memory, we can enable {\em (1)}~systematic \textit{exploration} of reasoning paths, {\em (2)}~\textit{atomic} subqueries enabling precise retrievals and grounding, and {\em (3)}~\textit{efficiency} through parallel execution and bounded context window utilization. Moreover, \ours's modular design allows it to be integrated with existing RAG methods, thus providing a practical solution to improve current RAG systems. On standard multi-hop reasoning benchmarks, \ours consistently achieves improvements over recently proposed methods such as RQ-RAG and Self-RAG, while maintaining comparable computational costs. 

\end{abstract}

\section{Introduction}
Retrieval-Augmented Generation \citep[RAG, ][]{lewis2020retrieval, petroni2020how, guu2020retrieval} has emerged as a promising approach for grounding language model (LM) responses in external knowledge. However, RAG systems struggle with multi-hop queries that require reasoning across multiple retrieved documents \citep{tang2024multihoprag, wei2022chain}. A key challenge lies in the initial retrieval step, which often fails to retrieve sufficient relevant documents due to the query’s lack of full contextual information \citep{ma2023query}. This limitation has been highlighted in recent surveys \citep{torfi2020natural, zhao2023survey} as a fundamental barrier to reliable AI systems, particularly given the widespread deployment of large language models \citep{brown2020language} across critical domains.
Consider the query: {\em ``Rumble Fish was a novel by the author of the coming-of-age novel published in what year by Viking Press?''} Answering this requires an iterative retrieval process: identifying the Rumble Fish's author, connecting to their coming-of-age novel, and determining its publication year. Single-step retrieval in RAG systems often fails in such cases, as it may retrieve documents about Rumble Fish's author and Viking Press without recognizing the intermediate fact---the author's coming-of-age novel---must first be established. Furthermore, \citet{leng2024long, shuster2021retrieval} demonstrate that even when relevant documents are retrieved, LMs struggle to reason across them due to fixed context windows, leading to information loss and broken reasoning chain. These limitations pose risks in critical domains such as healthcare and finance \citep{pal2023med, zhao2024revolutionizing}, where accurate multi-step reasoning is essential.

\begin{figure}[!t]
	\centering
	\input{fig/teaser.tex}\\[-1em]
	\caption{\textbf{\ours} improves performance on the HotpotQA benchmark substantially compared to various existing RAG methods, demonstrating the value of externalizing planning as a directed acyclic graph (DAG) outside of the LLM's context.}
	\label{fig:teaser_bar}
    \vspace*{-1em}
\end{figure}

\begin{figure*}[!t]
	\centering
	\small
	\begin{tikzpicture}[
		node distance=0.6cm and 0.0cm,
		base_style/.style={
			rectangle, 
			rounded corners, 
			minimum width=#1, 
			minimum height=.5cm, 
			align=center, 
			draw=black, 
			line width=.75pt,
			font=\small,
			fill=highlight,
			text width=#1
		},
		q_style/.style={ 
			minimum width=#1, 
			minimum height=.4cm, 
			align=left, 
			font=\small,
			text width=#1
		},
		arrow_style/.style={->, line width=1.0pt, draw=black!70},
		feature_arrow/.style={->, line width=1.2pt, draw=red!70, dashed},
		benefit_box/.style={
			draw,
			rounded corners,
			fill=white,
			text width=4.0cm,
			inner sep=0.3cm
		},
	prediction_style/.style={
		font=\footnotesize,
		text width=3cm,
		align=left
	}
		]
		
		\node (Q) [q_style=5cm] {};
		\node (dummy) [q_style=7.5cm] at ($(Q)+(.25,0)$) {\texttt{$Q$: Rumble Fish was a novel by the author of the coming-of-age novel published in what year by Viking Press?}};
		\node (Q11) [base_style=3.5cm, below=of Q,xshift=-1.5cm] {$Q1.1$ Who is the \mbox{author} of Rumble Fish?};
		
		\node (Q21) [base_style=3.5cm, below=of Q11] {$Q2.1$ What is the coming-of-age novel by $\langle$A1.1$\rangle$?};
		
		\node (Q31) [base_style=3.5cm, below=of Q21] {$Q3.1$ In what year was $\langle$A2.1$\rangle$ published by Viking Press?};
		
		\draw[arrow_style] (Q11) -- (Q21);
		\draw[arrow_style] (Q21) -- (Q31);
		
		\node[benefit_box,anchor=north west] at ($(Q11.north east)+(.5cm,0)$) (benefits) {
			\textbf{Existing RAG Methods:}
			\\
			\xmark~Implicit Reasoning \\
			\xmark~Context Overflow \\
			\xmark~Sequential Dependency \\[1em]
			\textbf{\ours:}\\ 
			\cmark~Test-time Planning \\
			\cmark~Conditional Independence \\
			\cmark~Dynamic Adaptation, $\langle$AI.J$\rangle$
		};

		\node[right=4cm of Q, yshift=-.0cm, align=left, text width=6cm, inner sep=0pt, font=\footnotesize] (vanilla) {
			\textbf{Vanilla RAG}\\ $\quad\fllm: Q \times \mathbf{D} \rightarrow G$
		};
		
		\node[below=3pt of vanilla, align=left, text width=6cm, inner sep=0pt, font=\footnotesize] (cot) {
			\textbf{Chain-of-Thought}\\ $\quad\fllm: Q \times \mathbf{D} \times \textcolor{orange}{\bigoplus}_{i=1}^{n} t_i \rightarrow G$
		};
		
		\node[below=3pt of cot, align=left, text width=6cm, inner sep=0pt, font=\footnotesize] (react) {
			\textbf{Reason \& React (ReAct)}\\ $\quad\fllm: Q \times \textcolor{orange}{\bigcup}_{i=1}^n (t_i {\times} a_i {\times} o_i {\times} \mathbf{D}_i) \rightarrow G$ 
		};
		
		\node[below=3pt of react, align=left, text width=6cm, inner sep=0pt, font=\footnotesize] (qd) {
			\textbf{Query Decomposition}\\ $\quad\fllm: Q \rightarrow \{q_i\}_{i=1}^n$\\ 
			$\quad\fllm: Q \times \mathbf{D} \times \textcolor{orange}{\bigcup}_{i=1}^n(q_i, G_i) \rightarrow G$ 
		};
		
		\node[below=3pt of qd, align=left, text width=6cm, inner sep=0pt, font=\footnotesize] (selfrag) {
			\textbf{Self-RAG}\\ $\quad\fllm : Q \times \mathbf{D} \rightarrow \{(G_i, S_i)\}_{i=0}^K$\\ 
			$\quad G = \textcolor{orange}{\arg\max}_{G_i} S_i$ 
		};
		
		\node[below=3pt of selfrag, align=left, text width=6cm, inner sep=0pt, font=\footnotesize] (ours) {
			\textbf{\ours}\\ $\quad \fllm: Q \rightarrow \textcolor{teal}{\mathcal{G}(\mathbf{V},\mathbf{E})}$\\ 
			$\quad \fllm: G(\textcolor{teal}{\mathbf{Pa}(q)}) \times q \times \mathbf{D} \rightarrow G(q) \quad \forall q \in \mathbf{V}$ 
		};

		\node[prediction_style, right=8.6cm of Q, yshift=.5cm] (pred1) {\texttt{Answer: 1967}};
		\node[prediction_style, right=0.2cm of vanilla,font=\large] (pred1) {(\xmark)};
		\node[prediction_style, right=0.2cm of cot,font=\large] (pred2) {(\xmark)};
		\node[prediction_style, right=0.2cm of react,font=\large] (pred3) {(\xmark)};
		\node[prediction_style, right=0.2cm of qd,font=\large] (pred4) {(\xmark)};
		\node[prediction_style, right=0.2cm of selfrag,font=\large] (pred5) {(\xmark)};
		\node[prediction_style, right=0.2cm of ours,font=\large] (pred6) {(\cmark)};
		
	\end{tikzpicture}\\[-8pt]
	\caption{\textbf{Comparison of RAG approaches:} Comparison of different RAG approaches on a HotpotQA multi-hop query. Traditional RAG methods (right) struggle with context management and implicit reasoning, where \textcolor{orange}{sequential operators} create implicit dependencies. In contrast, \ours (left) generates an explicit reasoning plan as a DAG (\textcolor{teal}{$\mathcal{G}$}) at \textit{test-time}, with special tags $\langle$AI,J$\rangle$ enabling dynamic information flow through parent subqueries (\textcolor{teal}{$\mathbf{Pa}(q)$}). On this example query, while previous approaches fail to identify the correct publication year, \ours successfully decomposes the reasoning process and arrives at the correct answer (1967).}
	\label{fig:method_comparison}
	\vspace*{-.5em}
\end{figure*}

Recent research has attempted to address these limitations through structured reasoning frameworks. Chain-of-Thought (CoT) prompting \citep{wei2022chain} and systematic query decomposition \citep{patel2022question} have introduced explicit reasoning steps, enabling more granular thought processes and targeted retrievals. Building upon these foundations, \citet{yao2023react} proposed ReAct---a framework that creates a reasoning chain inside the LM's context and dynamically coordinates between thoughts, retrieval operations (actions), and information processing (observations). However, a fundamental limitation of ReAct is that the reasoning and retrieval plan exists entirely within the LLM's context window, leading to context overflow as reasoning chains grow, plan fragmentation, and high latency from sequential execution. While recently proposed frameworks like Self-RAG \citep{asai2023self}, RA-ISF \citep{liu2024ra}, and RQ-RAG \citep{chan2024rqrag} have advanced the field through innovations in dynamic reasoning and adaptive retrieval mechanisms, the core problem still remains. These methods often struggle to balance reasoning depth and computational efficiency (see \cref{sec:motivation} for an example). This trade-off results in either shallow analysis that misses critical reasoning steps, or impractical latency that limits their deployment in real-time applications.

To address these challenges, we propose \ours, a novel approach that externalizes the reasoning process as a directed acyclic graph (DAG). Unlike existing methods that maintain reasoning chains within the LM's context or generate isolated sequential sub-queries, \ours decomposes the main query into interrelated \textit{atomic} and \textit{dynamic} sub-queries, where nodes represent atomic reasoning steps and edges capture the conditional dependencies between them. Atomic sub-queries can be answered by a single retrieval, and further decomposition does not provide additional useful granularity. Additionally, sub-queries are tagged with special tags $\langle$AI.J$\rangle$, allowing them to adapt based on updated knowledge during generation. This external representation addresses core limitations of existing approaches in several ways. First, by following the DAG structure for generation, only the parent nodes need to be included in the context window for any given step (and that the system follows the intended plan). This prevents context overflow and ensures that only relevant information is processed at each stage. Second, nodes at the same depth in the DAG can be processed in parallel, significantly reducing latency compared to sequential approaches. Third, nodes are dynamic, containing tags that allow them to adapt based on their parent's generation, resulting in more targeted sub-queries. Furthermore, the DAG structure enables explicit verification at each node, facilitating error correction and backtracking. As generation involves traversing through the reasoning DAG, \ours can be incorporated with both traditional RAG frameworks and recent approaches like Self-RAG \citep{asai2023self}, enhancing their multi-hop reasoning capabilities (\cref{fig:teaser_bar}) while maintaining comparable computational efficiency in terms of input-output tokens (\cf \cref{fig:plan_analysis}).

\textbf{Contributions}~~We summarize our contributions as follows.
{\em (i)}~We introduce a \textit{test-time} reasoning approach that externalizes the reasoning process as a DAG, fundamentally altering how multi-hop queries are processed in RAG systems.
{\em (ii)}~We propose a novel DAG-based reasoning structure that decomposes complex queries into \textit{atomic} and \textit{dynamic} sub-queries, enabling parallel processing while maintaining conditional dependencies.
{\em (iii)}~We demonstrate that our approach can enhance both traditional RAG frameworks and modern approaches like Self-RAG, improving multi-hop reasoning capabilities with comparable computation.\looseness-2

\section{Related Work}
\label{sec:related_work}
We review two key areas relevant to our work: {\em (1)}~reasoning in LLMs, which focuses on methods that enhance multi-step reasoning, and {\em (2)}~retrieval-augmented generation (RAG), which integrates external information to improve LLM performance in knowledge-intensive tasks.

\paragraph{Reasoning in LLMs} 
Recent advancements in LLMs have significantly enhanced their reasoning capabilities through various approaches. Chain-of-Thought (CoT) prompting \citep{wei2022chain} improves model performance by guiding LLMs through intermediate reasoning steps. Building on this, the Tree of Thoughts (ToT) framework \citep{yao2024tree} enables LLMs to explore and evaluate multiple reasoning paths simultaneously. \citet{wang2023selfconsistency} proposed self-consistency, a technique that samples multiple reasoning chains and selects the most likely answer through majority voting. Similarly, least-to-most prompting \citep{zhou2023leasttomost} decomposes complex questions into simpler subquestions, addressing them sequentially. Recently, \citet{hao2023reasoning} propose Reasoning via Planning (RAP), where LLMs act as both world models and reasoning agents. Additionally, \citet{sun2024thinkongraph} introduced Think-on-Graph (ToG), incorporating knowledge graphs into multi-hop reasoning for deeper and more interpretable reasoning processes. Recent works from \citet{welleck2024decoding, chen2024tree} also analyzed tree search for reasoning. While these methods have demonstrated impressive results, they still rely on implicit LM reasoning capability or on sequential reasoning steps, which can lead to increased context usage and computational costs. In contrast, \ours externalizes the reasoning process as a DAG, mitigating these inefficiencies.

\paragraph{Retrieval Augmented Generation} 
RAG enhances large language models (LLMs) by integrating relevant external documents, leading to notable performance improvements, particularly in knowledge-intensive tasks \citep{lewis2020retrieval, guu2020retrieval}. Retrieval strategies in RAG models can be categorized into three paradigms based on the frequency of retrievals: {\em(1)}~one-time retrieval, {\em(2)}~retrieval every $k$ tokens, and {\em(3)}~adaptive retrieval. Models employing one-time retrieval include DrQA \citep{chen2017reading}, REALM \citep{guu2020retrieval}, and ATLAS \citep{izacard2023atlas}. Retrieval at fixed intervals (every $k$ tokens) is used by RALM \citep{ram2023context}, RETRO \citep{pmlr-v162-borgeaud22a}, and InstructRetro \citep{pmlr-v235-wang24bd}. In contrast, adaptive retrieval approaches---such as Self-RAG \citep{asai2023self}, SPALM \citep{yogatama2021adaptive}, Adaptive kNN \citep{drozdov2022you}, and Active-Retriever \citep{jiang2023active}---dynamically adjust the frequency and nature of document retrieval based on task requirements and input context. FLARE \citep{jiang2023active} uses token probability distributions to trigger retrievals, while RETRO \citep{pmlr-v162-borgeaud22a} employs a specialized architecture for fixed-interval document retrieval. \citet{wang2024rat} proposed RAFT, combining chain-of-thought reasoning with RAG through iterative thought refinement. Other approaches include RA-ISF \citep{liu2024ra}, which combines dataset-specific specialized models, and RQ-RAG \citep{chan2024rqrag}, which enables query rewriting and decomposition for handling complex queries. \citet{hsu2024groundingtryingllmsreinforcement} also focus on multi-hop query performance of RAG but take an orthogonal approach.

Recent work by \citet{ranaldi2024eliciting}, explores reasoning in RAGs through C-RAG which generates explanations explicitly contrasting the relevance of retrieved passages to support the final answer. \ours takes a different approach by proposing a novel framework that performs reasoning at test-time outside the LM's working memory, and performs generation in a computationally efficient manner, enabling LLMs to tackle complex multi-hop queries with greater accuracy while producing a verifiable reasoning trace.\looseness-1

\section{Motivation}
\label{sec:motivation}
Multi-hop reasoning queries, requiring information from multiple documents, pose significant challenges for current Retrieval-Augmented Generation (RAG) systems. Consider an example query from HotPotQA \citep{yang2018hotpotqa}: \textit{``Rumble Fish was a novel by the author of the coming-of-age novel published in what year by Viking Press?"}. The query demands multiple reasoning steps: identifying the Rumble Fish's author, connecting to their coming-of-age novel, and determining its publication year. In this section, we will use this query as a running example {\em a)}~to articulate the failure modes of standard as well as recently-proposed RAG approaches on such multi-hop reasoning queries (\cref{fig:method_comparison}); and {\em b)}~motivate the core idea of our proposed solution. For this analysis, we fix the model to be Llama-3.1-instruct\textsubscript{8B} and the retriever to be Contriever \citep{izacard2022unsupervised}. 
In the remainder of the paper, $Q$ denotes queries, $D$ denotes documents, and $G$ denotes generations (responses). %

\paragraph{Standard Baselines} A straight-forward RAG approach is to issue the given query $Q$ to a retriever, retrieve documents $D$, and prompt a language model (LM) with $Q$ and $D$ to generate an answer. For the running example, this approach produces an incorrect answer (\textit{1975}), because {\em a)}~the retrievals are incomplete; it fails to get the ground-truth correct document on the coming-of-age novel \textit{The Outsiders}, which is not surprising as the query makes only an indirect reference to the novel, and {\em b)}~in turn, LM generates an incorrect answer without any reasoning.

This motivates using better reasoning at inference time. A natural idea is to prompt the same LM with $Q$ and $D$, but also elicit a reasoning for the answer in a chain-of-thought (CoT) manner \citep{wei2022chain}. 
However, for our example, RAG with CoT generates incorrect reasoning: \textit{[S.E.\ Hinton wrote Rumble Fish → Rumble Fish was published in 1975 → The author of The Outsiders also published in 1975]}, leading to the incorrect answer \textit{1975}. Here, we see the reason for failure is that the model hallucinated (third step of the chain) as it was trying to break down the complex query. %
Thus, the multi-hop query requires careful reasoning. Next, we turn to approaches that use more \textit{test-time} compute. %

\paragraph{Reason \& Act (ReAct)} ReAct \citep{yao2023react} extends traditional RAG frameworks by implementing a structured interaction loop between reasoning and retrieval. The framework decomposes each iteration into three components: a \textit{Thought} phase for planning, an \textit{Action} phase for doing retrievals, and an \textit{Observation} phase for incorporating retrieved information. For our example, ReAct produces the following plan (and retrievals): \textit{(T1) Need to find the author of the coming-of-age novel, then find its Viking Press publication year → $\langle$ \emph{Retrieve} $\rangle$ → (T2) Found S.D.\ Smith, need to verify Viking Press publication → $\langle$ \emph{Retrieve} $\rangle$ → (T3) Located publication info for Rumble Fish}, ultimately answering \textit{1975}. This trace reveals how ReAct, despite using more test-time compute, fails to keep track of the original query intent. ReAct approach relies entirely on the LM's in-context memory, that keeps growing with iterations and retrievals. It becomes challenging for the LM to track and update the plan periodically in light of the new retrievals.
As shown in \cref{fig:method_comparison}, while ReAct enables dynamic reasoning, it is vulnerable to error propagation, and struggles to maintain entity relationships across steps.

\paragraph{Query Decomposition (QD)} Query decomposition methods \citep{patel2022question} address the limitations of previous approaches by breaking queries into independent, sequential subqueries. LM first decomposes the main query into subqueries, then solves them sequentially while maintaining previous subquery-answer pairs in context. 
For our example, it generates two subqueries: \textit{``What is the author of the coming-of-age novel published by Viking Press?"} yielding \textit{``S.E. Hinton"}, followed by \textit{``In what year was the coming-of-age novel published by Viking Press?"} producing \textit{``1975"}. Firstly, we note that the decomposition is not very helpful. Secondly, the LM answers the subqueries based on the first document retrieved for the original query, which is on \textit{Rumble Fish} (and gets the author name right for the first sub-query). Despite explicitly decomposing the query, the method still fails to maintain coherent reasoning across subqueries, and loses critical entity relationships between subsequent sub-queries. Recent approaches such as RQ-RAG~\citep{chan2024rqrag} and RA-ISF~\citep{liu2024ra} extend query decomposition, and incorporate query rewriting tricks, but inherit similar limitations.

\paragraph{Self-RAG} \citet{asai2023self} recently proposed Self-RAG that introduces a novel self-evaluation mechanism. The LM generates multiple candidate responses and scores them based on their faithfulness to retrieved evidence as well as relevance to the original query. This approach aims to improve reasoning quality through verification. However, Self-RAG still fundamentally relies on the language model's implicit reasoning capabilities within a single context window, similar to the previously discussed approaches. When tested on our example query, it exhibits the same limitations, producing the incorrect answer \textit{1975} despite its sophisticated verification mechanism.

\begin{figure}[!t]
	\centering
	\resizebox{\columnwidth}{!}{\input{fig/reasoning_dag_example.tex}}\\[-4pt]
	\caption{\textbf{Reasoning plan example:} A Reasoning DAG generated by the reasoning plan expert, highlighting key advantages: only relevant information flows to each subquery, subqueries on the same depth can be executed in parallel, and the DAG structure allows for debugging and backtracking.}
	\label{fig:reasoning_dag_example}
\end{figure}

\begin{table*}[t!]
	\centering
	\caption{\textbf{Reasoning DAG depth} (percentage/count) for multi-hop (HotpotQA, StrategyQA, MuSiQue) and single-hop  (PopQA) data sets. For single-hop queries, the DAG primarily has depth 0, which is desirable, while multi-hop queries require deeper reasoning paths.\looseness-1}
	\setlength{\tabcolsep}{13pt}%
	\small	
	\begin{tabularx}{\textwidth}{lccccc}
		\toprule
		\textbf{Dataset} & \textbf{Depth 0} & \textbf{Depth 1} & \textbf{Depth 2} & \textbf{Depth 3} & \textbf{Depth$\geq$4} \\ \midrule
		\textbf{HotpotQA} {(Multi-hop)}    
		& \cellcolor{red!01}0.5\% (35)  
		& \cellcolor{red!13}12.8\% (945)  
		& \cellcolor{red!80}79.5\% (5884)  
		& \cellcolor{red!07}6.8\% (507)  
		& \cellcolor{red!0}0.4\% (34)  \\ 
		
		\textbf{StrategyQA} {(Multi-hop)}  
		& \cellcolor{red!01}0.9\% (22)   
		& \cellcolor{red!43}42.9\% (978)  
		& \cellcolor{red!51}51.2\% (1175)  
		& \cellcolor{red!5}4.5\% (103)  
		& \cellcolor{red!0}0.3\% (6)   \\ 
		
		\textbf{MuSiQue} {(Multi-hop)} 
		& \cellcolor{red!0}0.0\% (0)   
		& \cellcolor{red!2}2.11\% (51)  
		& \cellcolor{red!66}66.4\% (1604)   
		& \cellcolor{red!25}25.5\% (617)   
		& \cellcolor{red!5}5.9\% (145)   \\ 
		
		\textbf{PopQA} {(Single-hop)}      
		& \cellcolor{red!78}77.9\% (1090) 
		& \cellcolor{red!01}0.8\% (11)   
		& \cellcolor{red!19}18.9\% (264)   
		& \cellcolor{red!2}2.4\% (34)   
		& \cellcolor{red!0}0.0\% (0)   \\ 
		
		\bottomrule
	\end{tabularx}	
	\label{tbl:reasoning_depth}
	\vspace*{-.5em}
\end{table*}

The aforementioned methods incorporate different strategies and varying degrees of test-time compute to handle reasoning in RAG systems. But they all share fundamental limitations, listed below, that lead to failures even on 2--3 hop queries such as our running example. We identify two observations that motivate our proposed framework \ours:\looseness-1
\begin{itemize}[leftmargin=*,itemsep=0pt,topsep=2pt]
  \item \textbf{First}, by accumulating intermediate reasoning steps in the context window, they create information overload as the context grows. It becomes challenging for the LM to identify and resolve the dependencies between subqueries and faithfully execute a reasoning plan.
  \item \textbf{Second}, state-of-the-art RAG approaches fail to leverage the inherent independence between different reasoning paths in multi-hop queries. For instance, finding an author's identity and locating a publication date can be answered independently, without overburdening the LM's context with all reasoning paths. 
\end{itemize}

\section{\ours: Test-Time Planning}
\label{sec:proposed_framework}
We present \ours, a novel framework for multi-hop reasoning through \textit{dynamic, test-time} planning via Directed Acyclic Graphs (DAGs). The key ideas of our formulation, motivated by the issues faced by the current RAG systems discussed in \cref{sec:motivation}, are:
\begin{enumerate}[leftmargin=*,itemsep=0pt,topsep=2pt]
    \item We want to isolate the reasoning plan, outside of the LM's in-context memory, as a global data structure that can help with tracking, execution, and verification. We can generate the plan once statically, refine, and execute the plan using data structure operations. In this work, we use Directed Acyclic Graphs (\cref{fig:reasoning_dag_example}) that are appropriate to represent complex plans. 
    \item DAG structure allows executing any node conditioned only on the necessary context along its path, unlike state-of-the-art approaches that accumulate the entire trace (plan and retrievals) in the LM's in-context memory. This also helps make the LM calls more economical in terms of the number of tokens.
    \item DAG structure allows processing nodes independently, conditioned on the execution of parent nodes, helping with efficiency and end-to-end latency of RAG system.
\end{enumerate}

At a high level, \ours operates in two phases: {\em (1)}~decomposing complex queries into a DAG structure, and {\em (2)}~executing the DAG while preserving dependencies. Besides the merits of serializing the query plan as a data structure---such as control and efficiency listed above, \ours is complementary to state-of-the-art approaches like Self-RAG in the following sense. We can seamlessly integrate \ours with methods like Self-RAG, to solve the sub-queries in its internal nodes (see \cref{sec:exp_accuracy}).

We formalize the reasoning plan DAG in \cref{sec:proposed_dag}, followed by detailed analysis on properties and benefits of \ours in \cref{sec:plan_properties}. The full algorithm is showcased in \cref{alg:plan_rag}.

\subsection{Reasoning Plan: Directed Acyclic Graph}
\label{sec:proposed_dag}
At test-time, \ours generates a reasoning plan as a Directed Acyclic Graph (DAG) $\mathcal{G}(\mathbf{V},\mathbf{E})$, where $\mathbf{V}$ represents the set of generated subqueries and $\mathbf{E}$ denotes the directed edges between them. The root nodes of $\mathcal{G}$ correspond to independently answerable atomic subqueries, while subsequent nodes represent dependent queries that build upon their parent nodes' answers. This hierarchical structure follows the Markov assumption, ensuring that each node's answer depends only on its direct predecessors, thus enabling efficient parallel computation.

Formally, for any subquery $q {\in} \mathbf{V}$, its answer is computed as:\looseness-1
\begin{equation}
	G(q) = \fllm(G(\mathbf{Pa}(q)), q, \mathbf{D}_q) \, ; \quad \mathcal{G} = \fllm(Q) \, ,
\end{equation}
where $\mathbf{Pa}(q)$ denotes the set of parent nodes of $q$ in the DAG, $\mathbf{D}_q$ represents the retrieved documents relevant to $q$, and $Q$ is the main query. The language model $\fllm$ generates responses by jointly considering the subquery, its parent nodes' answers, and retrieved documents. When applied recursively, this formulation enables systematic reasoning from root nodes to leaves, with document retrieval potentially interleaved at each step.

To facilitate dynamic reasoning, we introduce a simple node indexing scheme. Each node is uniquely identified as $\langle i.j\rangle$, where $i$ denotes the node's depth from the root and $j$ indicates its position among nodes at depth $i$. Furthermore, we introduce a special tag \textbf{$\langle$AI.J$\rangle$} that enables dynamic dependency tracking between subqueries. In this notation, \textbf{I} and \textbf{J} are integer values representing the Question IDs required to complete a subquery. For instance, as illustrated in \cref{fig:reasoning_dag_example}, when subquery \textit{Q2.1} depends on the answer to \textit{Q1.1}, the tag $\langle$A1.1$\rangle$ enables dynamic answer propagation at inference time, allowing the system to adapt to updated knowledge.

\subsection{Generating a Reasoning Plan as a DAG}
\label{sec:plan_generation}
The first step during inference involves generating a reasoning plan by prompting the LLM with a specialized prompt, as detailed in \cref{app:reasoning_prompt}. While this initial plan is static, it can be dynamically instantiated, and refined, given the indexing scheme $\langle$AI.J$\rangle$ as discussed in \cref{sec:proposed_dag}. The plan therefore is a template initially which during generation dynamically materializes as the LM traverses through the DAG.

\begin{table}[t!]
	\centering\small
	\caption{\textbf{Evaluation with standard prompting methods:} Performance comparison of \ours against traditional RAG approaches (Vanilla-RAG, CoT-RAG, QD-RAG) and vanilla LLMs across different model scales.}
	\vspace*{3pt}
	\label{tbl:exp_simple_methods_accuracy}
	\setlength{\tabcolsep}{2.5pt}
	\input{fig/tbl_simple_acc.tex}
	\vspace*{-.5em}
\end{table}

To ensure reliable plan generation, we explore two approaches: \emph{(1)} fine-tuning a language model specifically for this structured output, and  \emph{(2)} leveraging more capable language models that can follow complex prompting instructions. Our experiments show both approaches are effective, with a fine-tuned Llama3.1-instruct\textsubscript{8B} achieving comparable performance to GPT-4o (\cref{sec:exp_reasoning_plan}). 

\subsection{Reasoning Plan: Properties \& Benefits}
\label{sec:plan_properties}

The effectiveness of \ours stems from its structured reasoning process, which offers several key advantages over traditional RAG-based methods. By leveraging a Directed Acyclic Graph (DAG) to represent reasoning plans, \ours ensures efficient information flow, reduces computational overhead, and improves query formulation. The DAG structure naturally enables systematic verification and granular control at the subquery level---errors can be isolated to specific nodes and automated interventions (like additional retrievals or alternative decompositions) can be targeted at specific reasoning steps. While we leave the full exploration of these verification and intervention capabilities to future work, this structural advantage distinguishes our approach from sequential reasoning methods. Below, we analyze the fundamental properties that enhance multi-hop reasoning while maintaining computational efficiency.

\paragraph{Atomic Subqueries} 
\ours decomposes complex queries into \textit{atomic} subqueries---unlike existing query decomposition methods that perform sequential, local decompositions. An atomic refers to subqueries that request a single piece of information and thus can be answered by a single retrieval---further decomposition does not yield additional useful granularity. This atomic nature enables precise retrievals (\cref{tbl:precision_recall}) and reduces hallucination risk by constraining the scope of each reasoning step. For instance, consider the query in \cref{fig:method_comparison}, instead of broad queries like \textit{``In what year was the coming-of-age novel published by Viking Press?"} (like query decomposition method), \ours generates specific atomic queries like \textit{``In what year was The Outsiders published by Viking Press?"} for \textit{Q3.1}, where the intermediate generation for $\langle$A2.1$\rangle$ is dynamically replaced with the corresponding generation.\looseness-1

\begin{algorithm}[t!]
	\caption{\ours framework}
	\label{alg:plan_rag}
	\begin{algorithmic}
		\footnotesize
		\STATE \textit{Input:} $Q$: Query
		\STATE \textit{Output:} Generation $G$
		\STATE Get a reasoning plan: $\mathcal{G} \gets \fllm(Q)$ 
		\STATE Identify root nodes of $\mathcal{G}$: $\mathbf{q}_{root}$
		\STATE Calculate depth of each node from root: \\ $l_q \gets maxdist(q, \mathbf{q}_{root})$
		\FOR {i: 0 to $\max_q (l_q)$}
		\FOR {parallel: $q$ in  $\{q: l_q = i \}$}
		\STATE Get parent questions and answers: 
		\STATE $M_{q} \gets \mathbf{Pa}(q)$ \& $M_{a} \gets G_{M_{q}}$
		\STATE $\tilde{q} \gets \fllm(q, M_{q}, M_a)$ \COMMENT{Generate the subquery}
		\STATE $G_q \gets  \fllm(\tilde{q})$ \COMMENT{Obtain generated answer for query $q$}
		\ENDFOR
		\ENDFOR
		\STATE $G \gets \Omega(\mathbf{q}, {G}_1, {G}_2, \dots, {G}_q)$  \COMMENT{Dataset Dependent Operator}
		\\
		\RETURN $G\phantom{G_q}$
	\end{algorithmic}
\end{algorithm}

\begin{figure*}[t]
	\centering
	\setlength{\figureheight}{.44\columnwidth}	
	\pgfplotsset{legend cell align={left},every axis/.append style={
			legend style={draw=none,inner xsep=2pt, inner ysep=0.5pt, nodes={inner sep=2pt, text depth=0.1em},fill=white,fill opacity=0.8}
	},xlabel style={yshift=5pt}}
	\setlength{\figurewidth}{.75\columnwidth}	
	\begin{subfigure}[b]{0.45\textwidth}
		\centering\footnotesize
		\input{fig/token_statistics.tex}%
	\end{subfigure}%
	\hfill
	\setlength{\figurewidth}{.82\columnwidth}	
	\begin{subfigure}[b]{0.52\textwidth}
		\centering\footnotesize
		\input{fig/hotpotqa_depth_comparison.tex}%
	\end{subfigure}\\[-1.2em]
	\caption{\textbf{Efficiency analysis:} (a)~Token utilization comparison showing that \ours maintains comparable computational efficiency with baseline methods. (b)~Analysis of reasoning depth on HotpotQA demonstrates that \ours naturally adapts to the dataset's 2-hop nature, achieving optimal depth for 80\% of queries while other methods show inconsistent reasoning depths.}
	\label{fig:plan_analysis}
    \vspace*{-3pt}
\end{figure*}

\paragraph{Parallelization \& Efficiency} 
The DAG structure explicitly captures conditional independence among subqueries through its edge relationships. This independence enables parallel execution of reasoning paths, significantly reducing latency compared to sequential approaches like query decomposition or ReAct. By controlling information flow through explicit parent-child relationships, \ours minimizes token overhead and improves computational efficiency (\cref{fig:plan_analysis}). Formally, for sequential methods,
\begin{equation}
	T^{\text{seq}} = \textstyle\sum_{i=1}^{n} (t_{\text{gen}}(q_i) + t_{\text{ret}}(\mathbf{D}_i)) \,,
\end{equation}
where $t_{\text{gen}}(q_i)$ is the generation time for subquery $q_i$ and $t_{\text{ret}}(\mathbf{D}_i)$ is the retrieval time for documents $\mathbf{D}_i$. In contrast, \ours enables parallel execution of independent nodes at the same depth, reducing the total computation time to:
\begin{equation}
	T^{\text{\ours}} = \textstyle\sum_{d=0}^{D} \max_{q \in \mathbf{V}_d} (t_{\text{gen}}(q) + t_{\text{ret}}(\mathbf{D}_q)) \,,
\end{equation}
where $\mathbf{V}_d$ represents the set of nodes at depth $d$ and $D$ is the maximum depth. Since, $T^{\text{\ours}} \leq T^{\text{seq}}$, the efficiency gain increases with the number of parallel branches. This parallelization significantly reduces latency compared to sequential approaches like query decomposition or ReAct.

\paragraph{Context Window Utilization} 
\ours achieves optimal context window utilization through selective information propagation. For sequential reasoning methods like Query Decomposition or ReAct, the context size grows linearly,
\begin{equation}
	C_t^{\text{seq}} = |Q| + \textstyle\sum_{i=1}^{t} (|q_i| + |G(q_i)| + |\mathbf{D}|) \, ,
\end{equation}
where $|Q|$ is the main query length, $q_i$ are intermediate questions, $G(q_i)$ their generations, and $\mathbf{D}_i$ the retrieved documents. In contrast, \ours maintains a constant context size at each node $q$,
\begin{equation}
	C_q^{\text{ours}} = |q| + |\mathbf{D}_q| + \textstyle\sum_{p \in \mathbf{Pa}(q)} (|p| + |G(p)|) \, .
\end{equation}
This selective propagation of only parents information, rather than entire reasoning history, enables efficient scaling to multi-hop queries while maintaining reasoning quality.

\section{Experiments}
We conduct extensive experiments to evaluate \ours across diverse reasoning scenarios, demonstrating significant improvements across multiple datasets.

\paragraph{Datasets} 
We evaluate \ours on four datasets that test different aspects of reasoning capabilities: HotpotQA \citep{yang2018hotpotqa} and StrategyQA \citep{Geva2021DidAU} for 2-hop reasoning, MuSiQue \citep{trivedi2022musique} for complex multi-hop reasoning (${>}2$ hops), and PopQA \citep{mallen2023llm_memorization} for single-hop queries. This diverse selection enables comprehensive evaluation of \ours's adaptability to varying reasoning depths. We consider an answer correct if the predicted answer contains the ground truth, providing a relaxed version of exact-match to account for variations in answer phrasing. Dataset details are provided in \cref{app:dataset_details}.

\paragraph{Retriever} 
For all experiments, we use the Contriever-MS MARCO \citep{izacard2022unsupervised} retriever with embeddings based on the 2018 English Wikipedia. The Wikipedia articles are segmented into non-overlapping 100-word chunks, and we retrieve the top-5 documents for each query.

\paragraph{Baselines and Methods} 
We evaluate \ours against several baselines as discussed in \cref{sec:motivation}: LM with no retrieval (Vanilla-LLM), standard retrieval with direct prompting (Vanilla-RAG), chain-of-thought prompting with retrievals (CoT-RAG), query decomposition with retrievals (QD-RAG), and state-of-the-art RAG methods, like Self-RAG \citep{asai2023self}, ReAct \citep{yao2023react}, and RQ-RAG \citep{chan2024rqrag}. These methods have outperformed methods like SAIL \citep{luo2023search}, Toolformer \citep{schick2024toolformer}, and commercial systems like Perplexity.ai and ChatGPT. We conduct experiments using three base models: GPT-3.5\textsubscript{turbo}, Llama2-chat\textsubscript{13B}, and Llama3.1-instruct\textsubscript{8B}.

We implement two variants of \ours: {\em (1)}~a base version that uses documents retrieved from the original query; {\em (2)}~\ours\textsubscript{SubQ} that performs iterative retrievals for each subquery in the DAG, enabling precise retrievals. Given \ours's agnostic design discussed in \cref{sec:proposed_framework}, we also evaluate its integration with Self-RAG~(Plan$^\ast$-Self-RAG).

\begin{table*}[t!]
	\centering
	\footnotesize
	\begin{minipage}[t]{0.57\textwidth}
		\captionof{table}{\textbf{Comparison with state-of-the-art methods:} Performance of \ours against recent advanced RAG methods (RQ-RAG, Self-RAG) and their Plan$^\ast$-augmented variants. \ours\textsubscript{SubQ} achieves consistent improvements across both single-hop and multi-hop reasoning tasks.}
		\footnotesize
		\setlength{\tabcolsep}{4.2pt}
		\renewcommand{\arraystretch}{1.28}
		\label{tbl:exp_sota_accuracy}

\input{fig/tbl_sota_acc.tex}
	\end{minipage}
	\hfill
	\begin{minipage}[t]{0.40\textwidth}
		\captionof{table}{\textbf{Retrieval effectiveness:} Evaluation of retrieval quality on HotpotQA distractor setting. \ours achieves higher precision and recall compared to ReAct and query decomposition.}
		\label{tbl:precision_recall}	
		\setlength{\tabcolsep}{5.7pt}
		\begin{tabularx}{\textwidth}{lccc}
			\toprule
			\textbf{Method} & \textbf{Precision} & \textbf{Recall} & \textbf{Accuracy} \\
			\midrule
			ReAct & $\phantom{0}0.04$ & $34.61$ & $25.68$ \\
			QD-RAG\textsubscript{SubQ}  & $\phantom{0}0.02$ & $32.16$ & $23.54$ \\
			\rowcolor{highlight}Plan$^\ast$RAG\textsubscript{SubQ} & $\mathbf{27.43}$ & $\mathbf{36.57}$ & $\mathbf{31.27}$ \\
			\bottomrule
		\end{tabularx}\\[-5pt]
		\captionof{table}{\textbf{Reasoning planners:} Accuracy comparison showing that a \textit{fine-tuned} Llama3.1-instruct\textsubscript{8B} planner achieves comparable performance to GPT-4o.}
		\label{tbl:llama3_plan_agent}
		\setlength{\tabcolsep}{2.5pt}
		\begin{tabularx}{\textwidth}{@{}ccccc@{}}
			\toprule
			\textbf{Vanilla-RAG} & 
			\textbf{CoT-RAG} & 
			\textbf{QD-RAG} & 
			\textbf{Plan\textsubscript{4o}} & 
			\textbf{Plan\textsubscript{8B}} \\
			\midrule
			$24.83$ & $30.11$ & $28.25$ & $31.97$ & $31.28$ \\
			\bottomrule
		\end{tabularx}
	\end{minipage}
\end{table*}

\subsection{Accuracy Performance}
\label{sec:exp_accuracy}
We evaluate \ours against both standard prompting techniques and state-of-the-art RAG frameworks in \cref{tbl:exp_simple_methods_accuracy} and \cref{tbl:exp_sota_accuracy}.
For fair comparison with standard prompting methods, all approaches use the same set of top-5 documents retrieved via Contriever-MS MARCO from the original query. As shown in \cref{tbl:exp_simple_methods_accuracy}, we compare against vanilla LLMs and three RAG variants (standard retrieval, chain-of-thought prompting, and query decomposition) using GPT-3.5\textsubscript{turbo}, Llama2-chat\textsubscript{13B}, and Llama3.1-instruct\textsubscript{8B}. \ours demonstrates consistent performance gains across all datasets and baseline models.

For state-of-the-art methods that employ iterative retrieval, we evaluate \ours\textsubscript{SubQ}. As shown in \cref{tbl:exp_sota_accuracy}, \ours\textsubscript{SubQ} achieves significant improvements over recent frameworks like Self-RAG, ReAct, and RQ-RAG across respective base models. Additionally, integrating Self-RAG to \ours's reasoning DAG (Plan$^\ast$-Self-RAG) yields substantial performance gains, validating the effectiveness of the \textit{test-time, external} planning approach.

These results establish that \ours's \textit{test-time} planning provides clear advantages over existing reasoning frameworks, including those designed for multi-hop reasoning.

\subsection{Reasoning DAG Depth} 
\cref{tbl:reasoning_depth} shows the reasoning DAG depths for all the datasets. As expected, the multi-hop query datasets exhibit a DAG depth greater than 1, indicating multiple atomic queries must be answered at different depths to answer the main query. In contrast, the single-hop dataset typically shows a reasoning depth of $0$, which is both expected and desired, as simpler queries do not require further decomposition into subqueries. This showcases that the reasoning DAG effectively adapts its complexity based on the query complexity.

On HotpotQA (a 2-hop dataset), we compare depth distributions for sequential methods like Query-Decomposition RAG (QD-RAG) and ReAct in \cref{fig:plan_analysis}. Both the methods exceed the ideal reasoning depth, with a significant proportion of samples having depths greater than 3, indicating suboptimal reasoning paths.

\paragraph{Information Gain (IG)}
We introduce \textit{cumulative information gain (IG)} to measure information aggregation across reasoning depths. Formally, at depth $d$,
\begin{equation}
	\label{eq:information_gain}
	IG(Q, \mathcal{G}, d) = IG(Q, \{q_i\}_{i=0}^d) \, ,
\end{equation} 
where $Q$ represents the main query and $\{q_i\}_{i=0}^d$ denotes the set of subqueries up to depth $d$. A higher IG indicates substantial new information contribution toward the final answer, while a similar value as the previous depth suggest redundant or unnecessary reasoning steps. 
We experiment with the HotpotQA dataset and employ GPT-3.5\textsubscript{turbo} model to score the information gain between 1-10 across the reasoning plan. \cref{tbl:info_gain_depth} showcases the average IG score over reasoning depth, demonstrating consistent incremental information gain as the DAG is traversed.

\subsection{Targeted Subqueries and Retrievals}
A key feature of \ours's reasoning plan is the \textit{atomic, dynamic} subqueries. As discussed in \cref{sec:plan_properties}, the subqueries in the reasoning plan are more targetted and help improve retrievals. We evaluate retrieval effectiveness with the HopotQA Distractor setting,  which provides ground truth labels for supporting documents. As shown in \cref{tbl:precision_recall}, \ours significantly outperforms both ReAct and Query-Decomposition (QD-RAG\textsubscript{SubQ}) in both precision and recall, contributing to improved accuracy.

High precision is particularly crucial for multi-hop reasoning, as it minimizes the introduction of irrelevant information that could propagate errors through the reasoning chain.

\subsection{Reasoning Plan Generation}
\label{sec:exp_reasoning_plan}
So far in the experiments we have utilized the more capable GPT-4o model for generating reasoning plans. However, to demonstrate that \ours framework is not dependent on powerful LLMs, we finetune a Llama3.1-instruct\textsubscript{8B} reasoning planner with LoRA adapters on a small subset of HotpotQA-train dataset. We use input-output pairs of questions and their corresponding reasoning DAGs in the required JSON format as training data. Experiments on a HotpotQA-dev subset (1300 samples) demonstrate comparable performance to GPT-4o, achieving [$31.28$] accuracy versus [$31.97$] with GPT-4o (\cref{tbl:llama3_plan_agent}). These results indicate that the success of our \textit{test-time} planning approach does not hinge on specific LLMs; and that modest training data is sufficient to fine-tune LMs to generate structured output.

\section{Discussion and Conclusion}
In this paper, we presented \ours, a framework that transforms multi-hop reasoning in RAG through \emph{test-time} planning. Our key insight---externalizing the reasoning structure as a DAG outside the LM's context---addresses fundamental limitations of existing approaches. Empirical results validate three core benefits: systematic exploration of reasoning paths, efficient parallel execution, and improved performance on multi-hop datasets. Significant improvements in retrieval precision demonstrates that atomic subqueries enable more focused document retrieval, while bounded context windows overcome the overflow challenges faced by sequential approaches. A key strength of \ours is its modular design, demonstrated by successful integration with existing RAG methods like Self-RAG. This flexibility, combined with comparable computational costs, establishes \ours as a practical solution for real-world multi-hop reasoning tasks

A key limitation in \ours is reliance on a static plan generated a priori. While the plan may need updates as more information is collected, the DAG structure facilitates subquery updates and backtracking since it is a formal object outside the LLM's context window. Future work can explore: {\em (1)}~verification and backtracking using the reasoning DAG, {\em (2)}~feedback loops between reasoning steps for dynamic plan refinement, and {\em (3)}~extending the framework to tasks like fact verification and mathematical deduction.

\clearpage

\bibliographystyle{icml2025}

\newpage
\appendix
\onecolumn

\section*{Appendices}
The supplementary document is organized as follows: \cref{app:dataset_details} presents the characteristics and specifications of datasets used in the evaluations. \cref{app:prompting_methods} details baseline models employing traditional prompting approaches, their implementation and specific prompts. \cref{app:sota_methods} describes the state-of-the-art RAG methods evaluated in the experiments. \cref{app:plan_rag_details} elaborates on the proposed \ours framework, encompassing reasoning plan generation, dynamic subquery generation, and answer generation through DAG traversal. Finally, \cref{app:experiment_details} provides details on the experiment setup.

\section{Dataset details}
\label{app:dataset_details}
In this section, we discuss the datasets used in the experiments. The datasets are particularly characterized into multi-hop and single-hop depending on the nature of the queries they contain.

\subsection{Multi-Hop QA} For multi-hop queries, we focus on three datasets: HotpotQA \citep{yang2018hotpotqa}, StrategyQA \citep{Geva2021DidAU}, and MuSiQue-Ans \citep{trivedi2022musique}

\paragraph{HotpotQA} HotpotQA is a multi-hop dataset collected from English Wikipedia. The questions are diverse and not constrained to any pre-existing knowledge bases or knowledge schemas. The dataset contains 7,405 queries in the \textit{dev-fullwiki} split. Although, each question in the dataset comes with two gold paragraphs, as well as a list of sentences in these paragraphs that crowd workers identify as supporting facts necessary to answer the question, we do not use these in our experiments and use contriever to fetch the relevant documents from the Wikipedia embeddings.

\paragraph{StrategyQA} StrategyQA is a question-answering benchmark requiring multiple reasoning steps to answer each question. Questions are short, topic-diverse, and require specific reasoning strategies for answers. We use the dataset that is available on the Self-RAG repository \citet{asai2023self}.  The dataset consists of 2,234 question-answer pairs. The dataset covers a broad range of topics including science, history, and common sense reasoning.

\paragraph{MuSiQue-Ans} MuSiQue-Ans is a multi-hop question answering dataset comprising 2,417 questions spanning 2 to 4 reasoning hops. Compared to existing QA datasets, MuSiQue-Ans presents a higher level of difficulty, evidenced by a threefold increase in the gap between human and machine performance. It is specifically designed to be resilient against models that rely on disconnected or shallow reasoning patterns.

\subsection{Single-Hop QA} To evaluate model performance on single-hop queries, we use PopQA \citep{mihaylov2018suit}.

\paragraph{PopQA} PopQA is an open-domain question-answering dataset designed to assess a model’s ability to retrieve and generate answers based on factual knowledge. The dataset consists of factual questions, many of which require specific knowledge of popular culture, history, and general world facts. In total the size of the dataset is 1,399 question-answer pairs. 

\section{Standard Prompting Methods}
\label{app:prompting_methods}
In this section we discuss various standard prompting methods experimented in \cref{sec:exp_accuracy} (\cf \cref{tbl:exp_simple_methods_accuracy}) along with the specific prompts. For all the models, we set temperature${=}0$ and as all the RAG methods discussed below use the query to retrieve top-k documents, all the methods use the same set of retrieved documents.

Note that for the Llama2 model, we wrap the prompts with the required system tags: \texttt{<s>,[INST],<<SYS>>,<</SYS>>}.

\paragraph{Vanilla-LLM} We experiment with three LLM models without any retrievals to answer the queries using its parametric knowledge: GPT-3.5\textsubscript{turbo}, Llama2-chat\textsubscript{13b}, and Llama3.1-instruct\textsubscript{8b}. Below is the prompt that we use:

\begin{lstlisting}[frame=single]
Be precise and give answer to the query. Response should be a valid JSON, that can be passed to json.loads directly, with a key as Response which only has 2-3 words. Do not use complete sentences or punctuation. In JSON, put every value as a string always, not float.

Example:

Query: What is the capital of France?
{"Response": "Paris"}
Query: How do you make coffee?
{"Response":  "Brew ground beans"}
\end{lstlisting}

\paragraph{Vanilla-RAG} We experiment with the standard RAG method, \ie retrieving top-k relevant documents for each query and providing them as context to the LLM. For our experiments, we use Contriever~\citep{izacard2022unsupervised} as the retriever with $k{=}5$. Similar to Vanilla-LLM, we experiment with 3 models: GPT-3.5\textsubscript{turbo}, Llama2-chat\textsubscript{13b}, and Llama3.1-instruct\textsubscript{8b}. Below is the prompt that we use:

\begin{lstlisting}[frame=single]
You are a concise answering assistant. Use the Retrievals while generating the answer and keep the answer grounded in the retrievals. Generate a JSON with a single key "Response" and a value that is a short phrase or a few words. In JSON, put every value as a string always, not float.
Note: Generate only JSON and no explanation or repeating the query or retrievals.

Example:

Query: Who was the PM of India when India performed its first nuclear test?
Retrievals: [["Indira Gandhi was the PM of India in 1974"], ["Indira Gandhi was the first woman PM of India."]]
Generation: {
	"Response": "Indira Gandhi"
}

Query: What is the capital of France?
Retrievals: [["Paris is the capital of France."]]
Generation: {
	"Response": "Paris"
}
\end{lstlisting}

\paragraph{CoT-RAG} We combine chain-of-thought (CoT) prompting with RAG, using Contriever as the retriever with $k{=}5$. We experiment with the three models: GPT-3.5\textsubscript{turbo}, Llama2-chat\textsubscript{13b}, and Llama3.1-instruct\textsubscript{8b}. Below is the prompt that we use:

\begin{lstlisting}[frame=single]
You are a highly intelligent assistant skilled at solving problems by reasoning step-by-step.  For each query, explain your reasoning clearly and logically in multiple steps, and then provide the final answer (Response). Use retrievals and keep your generation grounded in the retrievals. Produce a list of reasoning steps along with the Answer (Response) (precise and max 3-4 words). 
Note: Generate only reasoning steps and response. no explanation or repeating the query or retrievals. Strictly follow the below template and only return the JSON with no explanation.

Example:

Query: What is the sum of 15 and 27?
Retrievals: []
Generation: {
	"Reasoning_steps" : ["The first number is 15",  "The second number is 27", "Adding these gives 15 + 27 = 42"],
	"Response": "42"
}

Query: Who was the PM of India when India performed its first nuclear test?
Retrievals:[["Indira Gandhi was the PM of India in 1974"], ["Indira Gandhi was the first woman PM of India."]]
Generation:{
	"Reasoning_steps" : ["India performed its first nuclear test in 1974", "Indira Gandhi was the PM of India in 1974."],
	"Response": "Indira Gandhi"
}
\end{lstlisting}

\paragraph{QD-RAG} We experiment with query decomposition (QD) based RAG, where the original query is first broken down into simpler sub-queries. We use Contriever as the retriever with $k{=}5$ and evaluate three models: GPT-3.5\textsubscript{turbo}, Llama2-chat\textsubscript{13b}, and Llama3.1-instruct\textsubscript{8b}. Below is the prompt used for query decomposition:

\begin{lstlisting}[frame=single]
You are a helpful assistant that breaks complex queries into simpler ones that are easy to answer. Therefore, your job is to simplify complex queries into multiple queries that can be answered in isolation to eachother. If the query is simple, then keep it as it is.
NOTE: Always return a python list of subqueries that be passed to eval() directly.

Examples:

Query: Did Microsoft or Google make more money last year?
Subqueries: [`How much profit did Microsoft make last year?', `How much profit did Google make last year?']

Query: What is the capital of France?
Subqueries: [`What is the capital of France?']

Query: Who has the highest score in the last two ODI cricket world cups?
Subqueries: [`Who has the highest score in the last ODI cricket world cup?', `Who has the highest score in the second last ODI cricket world cup?', `Who scored the highest score among these two?']	
\end{lstlisting}
After decomposing the query into sub-queries, we solve each sub-query sequentially. The answers to previous sub-queries are provided as ``Known answers" to help maintain consistency and build upon intermediate findings. Below is the prompt used for answering each sub-query:

\begin{lstlisting}[frame=single]
You are a concise answering assistant. Use the Retrievals while generating the answer and keep the answer grounded in the retrievals. If Known answers are given, use them while generating the response. Generate a JSON with a single key "Response" and a value that is a short phrase or a few words. In JSON, put every value as a string always, not float.
Strictly follow the format below, and provide only the "Generation" part.

Example:

Query: Who was the PM of India when India performed its first nuclear test?
Retrievals: [["Indira Gandhi was the PM of India in 1974"], ["Indira Gandhi was the first woman PM of India."]]
Known answers:  Q=When did India perform the first nuclear test? A=India conducted its first nuclear test on May 18, 1974, at the Pokhran Test Range in Rajasthan, India.
Generation: {
	"Response": "Indira Gandhi was the PM of India when India performed its first nuclear test in 1974."
}

Query: What type of literature did John Keble write?
Retrievals: [["John Keble (1792-1866) was an English clergyman, poet, and theologian, best known for his contributions to religious poetry"], ["Keble wrote essays and sermons emphasizing the importance of tradition, the authority of the Church, and the significance of the sacraments."]]
Generation: {
	"Response": "Religious and devotional poetry"
}
\end{lstlisting}

\section{State-of-the-Art RAG Methods}
\label{app:sota_methods}
In this section we discuss various state-of-the-art RAG methods experimented in \cref{sec:exp_accuracy} (\cf \cref{tbl:exp_sota_accuracy}). 

\paragraph{Self-RAG} \citet{asai2023self} proposed Self-RAG, wherein the base Llama2 models are trained to learn a set of special reflection tokens. The reflection tokens are then used at the time of inference to judge the requirement for retrievals, relevance of the retrieved documents and the accuracy of the output. We tested Self-RAG\textsubscript{7b} and Self-RAG\textsubscript{13b} on both single and multi-hop datasets. As a retriever, we use contriever and set $k{=}5$. The temperature for both the models was set to 0 to maintain non-stochasticity. We use the official codebase released by the authors for all the experiments. 

\paragraph{RQ-RAG} \citet{chan2024rqrag} proposed RQ-RAG, a framework where the base Llama2 model is trained to enable it to dynamically refine search queries through rewriting, decomposing, and clarifying ambiguities. Further, control tokens are used to direct the generation process. In addition, the authors use three different sampling methods which includes selection based on perplexity (PPL), confidence, and an ensemble approach, in order to select the final answer. In our experiments, total of $k{=}5$ documents are retrieved at each depth for any given query and the maximum depth is set to $2$. The ensemble answer is used in all the experiments as that gave the best results. Similar to Self-RAG, temperature was set to $0$ to remove any stochasticity. We use the official codebase released by the authors for all the experiments. 

\paragraph{ReAct} ReAct \citep{yao2023react} combines reasoning and acting with language models to solve diverse language reasoning and decision-making tasks. The model uses an interleaved sequence of Thought, Action, and Observation steps to answer questions. For the experiment, we use Contreiver as a retriever, and the actions are limited to two options: \textit{`search'} and \textit{`finish'}. Therefore, the prompt for the models were changed accordingly. We experimented using both Llama3.1-instruct\textsubscript{8b} and GPT-3.5\textsubscript{turbo} models, with the temperature set to $0$ in both cases. We used only one retrieval ($k{=}1$) per thought as setting $k{=}5$ documents per thought gave poor results due to exploding contexts. The prompt used is as follows:

\begin{lstlisting}[frame=single]
You are a question answering agent, you need to solve the given question with interleaving Thought, Action, Observation steps. You need to do this one step at a time, given some previous steps. Thought can reason about the current situation, and Action can be of two types:
Search[entity], which searches and returns the top 1 relevant article. 
Finish[answer], which returns the answer and finishes the task.

Action can be only of the above two forms. They need to either include Search or Finish with the entity or answer enclosed in the bracket respectively.

Here are some examples:

Question: Musician and satirist Allie Goertz wrote a song about the "The Simpsons" character Milhouse, who Matt Groening named after who?

Thought 1: I only need to search Milhouse and find who it is named after.
Action 1: Search[Who was Milhouse named after?]
Observation 1: [Lisa kisses Milhouse. Lisa told Milhouse he should not give up searching for other girls and that life has unexpected things to offer. She told him he is cute in the moonlight, which caused him to fall off a cliff nearly to his death, but a bald eagle caught him, which left him saying "everything is coming up Milhouse!" Milhouse was designed by Matt Groening for a "Butterfinger" commercial, and it was decided to use the character in the series. Milhouse was named after U.S. President Richard Nixon, whose middle name was Milhous. The name was the most "unfortunate name"]

Thought 2: Milhouse was named after U.S. president Richard Nixon, so the answer is Richard Nixon.
Action 2: Finish[Richard Nixon]
\end{lstlisting}

\section{\ours}
\label{app:plan_rag_details}
We present the prompts and details of \ours, including its reasoning plan, tag replacement, and answer generation LM. For all the models, we set temperature${=}0$. Note that for the Llama2 model, we wrap the prompts with the required system tags: \texttt{<s> [INST] <<SYS>>} and \texttt{<</SYS>>}.

\subsection{Reasoning Plan}
\label{app:reasoning_prompt}
The reasoning plans, as discussed in \cref{sec:plan_generation}, are created by prompting the language model with the query and a set of contextual examples. The prompt guides the LLM to generate a minimal DAG where each sub-query's answer depends only on its parent nodes. We employ the special tag $\langle$AI.J$\rangle$ in sub-queries to explicitly denote dependencies on parent node answers. The prompt used for the DAG generation is:

\begin{lstlisting}[frame=single]
You are a reasoning DAG generator expert. The goal is to make a reasoning DAG with minimum nodes. Given a query, if it is complex and requires a reasoning plan, split it into smaller, independent, and individual subqueries. The query and subqueries are used to construct a rooted DAG so make sure there are NO cycles and all nodes are connected, there is only one leaf node with a single root and one sink. DAG incorporates Markov property i.e. you only need the answer of the parent to answer the subquery. The main query should be the parent node of the initial set of subatomic queries such that the DAG starts with it. Return a Python list of tuples of parent query and the subatomic query which can be directly given to eval().  

Strictly follow the below template for output.

For the subquery generation, input a tag <AI.J> where the answer of the parent query should come to make the query complete. 

NOTE: Make the DAG connected and for simple queries return the original query only without any reasoning DAG.

Example:

Query: Who is the current PM of India?
DAG: "Q: Who is the current PM of India?"

Query:What is the tallest mountain in the world and how tall is it?
DAG: [
	("Q: What is the tallest mountain in the world and how tall is it?", "Q1.1: What is the tallest mountain in the world?"), 
	("Q1.1: What is the tallest mountain in the world?", "Q2.1: How tall is <A1.1>?")
]

Query: What percentage of the worlds population lives in urban areas?
DAG: [
	("Q: What percentage of the worlds population lives in urban areas?", "Q1.1: What is the total world population?"), 
	("Q: What percentage of the worlds population lives in urban areas?",  "Q1.2: What is the total population living in urban areas worldwide?"), 
	("Q1.1: What is the total world population?", "Q2.1: Calculate the percentage living in urban areas worldwide when total population is <A1.1> and population living in urban areas is <A1.2>?"), 
	("Q1.2: What is the total population living in urban areas worldwide?", "Q2.1: Calculate the percentage living in urban areas worldwide when total population is <A1.1> and population living in urban areas is <A1.2>?")
]
\end{lstlisting}

\subsection{Dynamic Subquery Generation} The interdependence of child nodes on their parents is captured through special $\langle$AI.J$\rangle$ tags. During sub-query answer generation, we prompt the LLM to replace these tags with their corresponding parent answers to create coherent sub-query question. Below is the prompt used for tag replacement:

\begin{lstlisting}[frame=single]
You are provided a question with tags where the corresponding tag answers need to be replaced. Replace tags with answers of the previous question in such a way that the final question is coherent and logical.
Just replace all parts of the answers in the main question. Do not reason or answer the question. Your role is just to replace tags with all parts of the answer. 

NOTE: Only output the question with no explanation or any other details.

Example:

Query: Q2.1: Who was the president of India when the captain of the Indian cricket team was <A1.1> and vice-captain was <A1.2> in 2018?
Q1.1: Who was the captain of India cricket team in 2018?
A1.1: The captain of Indian cricket team in 2018 was M.S.Dhoni.
Q1.2: Who was the vice-captain of India cricket team in 2018?
A1.2: The vice-captain of Indian cricket team in 2018 was Virat Kohli.
Output: Q2.1: Who was the president of India when the captain of the Indian cricket team was M.S.Dhoni and vice-captain was Virat Kohli?
\end{lstlisting}

\subsection{Generator LLM} \ours involves traversing the reasoning plan to generate answers for each sub-query and ultimately resolve the main query. The following prompt is used for each answer generation:

\begin{lstlisting}[frame=single]
You are a concise answering assistant. If relevant and provided, use the Retrievals while generating the answer or use your own knowledge. If Known answers are given, use them while generating the response. Generate a JSON with a single key "Response" and a value that is a short phrase or a few words. In JSON, put every value as a string always, not float. Strictly follow the format below, and provide only the "Generation" part.

Example:

Query: Who was the PM of India when India performed its first nuclear test?
Retrievals: [["Indira Gandhi was the PM of India in 1974"], ["Indira Gandhi was the first woman PM of India."]]
Known answers: Q=When did India perform the first nuclear test? A=India conducted its first nuclear test on May 18, 1974, at the Pokhran Test Range in Rajasthan, India.
Generation: {
	"Response": "Indira Gandhi was the PM of India when India performed its first nuclear test in 1974."
}

Query: What type of literature did John Keble write?
Retrievals: [["John Keble (1792-1866) was an English clergyman, poet, and theologian, best known for his contributions to religious poetry"], ["Keble wrote essays and sermons emphasizing the importance of tradition, the authority of the Church, and the significance of the sacraments."]]
Generation: {
	"Response": "Religious and devotional poetry"
}
\end{lstlisting}

\section{Experiments}
\label{app:experiment_details}
This section presents comprehensive details of our experimental evaluations. We begin with an analysis of a HotpotQA example query (\cref{fig:method_comparison}), demonstrating comparative method outputs in \cref{app:hotpotqa_comparison}. We then evaluate performance on multi-hop and single-hop datasets in \cref{app:performance_experiment}, presenting results in \cref{tbl:exp_simple_methods_accuracy} and \cref{tbl:exp_sota_accuracy}. Following this, we analyze \ours's retrieval effectiveness in \cref{app:retrieval_precision} and information gain results in \cref{app:information_gain}. Finally, we describe the reasoning plan generation experiments using a finetuned Llama3.1-instruct\textsubscript{8b} model in \cref{app:reasoning_agent}.

\subsection{HotpotQA Example Query}
\label{app:hotpotqa_comparison}
We analyze a query from HotpotQA to demonstrate how different methods approach multi-hop reasoning:

\textbf{Query:} ``Rumble Fish was a novel by the author of the coming-of-age novel published in what year by Viking Press?"\\
\textbf{Ground Truth:} 1967

As discussed in \cref{sec:motivation}, we compare various methods using Llama3.1-instruct\textsubscript{8b} with temperature${=}0$ and Contriever as the retriever ($k{=}5$). Below are the outputs for each method:

\paragraph{Vanilla-LLM} generates \textit{S.E. Hinton 1975}.

\paragraph{Vanilla-RAG} generates \textit{1975}.

\paragraph{Chain-of-Thought (CoT) - RAG} produces the following reasoning chain:\\
\textit{S.E. Hinton wrote Rumble Fish} ${\rightarrow}$ \textit{Rumble Fish was published in 1975} ${\rightarrow}$ \textit{The author of The Outsiders also published in 1975}\\
Final generation: \textit{1975}

\paragraph{ReAct} generates the following sequence:\\
\textit{Thought 1:} I need to find the author of the coming-of-age novel, then find the year it was published by Viking Press.\\
\textit{Action 1:} Search[Author of the coming-of-age novel published by Viking Press]\\
\textit{Thought 2:} The author of the coming-of-age novel is S.D. Smith, but I am not sure if he published it with Viking Press.\\
\textit{Action 2:} Search[Coming-of-age novel by S.D. Smith published by Viking Press]\\
\textit{Thought 3:} I found that S.E. Hinton published Rumble Fish, and it was published by Viking Press in 1975.\\
\textit{Action 3:} Finish[1975]\\
Final generation: \textit{1975}

\paragraph{Query Decomposition (QD) - RAG} decomposes the query as follows, we show both the query and the generation:
\textit{Q1:} What is the author of the coming-of-age novel published by Viking Press?\\
\textit{A1:} S.E. Hinton\\
\textit{Q2:} In what year was the coming-of-age novel published by Viking Press?\\
\textit{A2:} 1975\\
\textit{Q3:} What is the title of the novel written by the author of the coming-of-age novel?\\
\textit{A3:} The Outsiders\\
Final generation: \textit{1975}

\paragraph{Self-RAG} using the fine-tuned Llama2\textsubscript{13b} model generates \textit{1975}.

\paragraph{\ours} generates the following reasoning plan:\\
\textit{Q1.1:} Who is the author of Rumble Fish? ${\rightarrow}$ \textit{Q2.1:} What is the coming-of-age novel by $\langle$A1.1$\rangle$? ${\rightarrow}$ \textit{Q3.1:} In what year was $\langle$A2.1$\rangle$ published by Viking Press?

During the generation the subquery and generation are as follows:\\
\textit{Q1.1:} Who is the author of Rumble Fish?\\
\textit{A1.1:} S. E. Hinton\\
\textit{Q2.1:} What is the coming-of-age novel by S. E. Hinton?\\
\textit{A2.1:} The Outsiders\\
\textit{Q3.1:} In what year was The Outsiders published by Viking Press?\\
\textit{A3.1:} 1967\\
Final generation: \textit{1967}

Therefore, \ours is the only method that arrives at the correct answer of \textit{1967}.

\subsection{Performance Experiment}
\label{app:performance_experiment}
\cref{tbl:exp_simple_methods_accuracy} and \cref{tbl:exp_sota_accuracy} demonstrate the performance limitations of traditional prompting RAG methods and recently proposed advanced RAG approaches, while highlighting the superior performance of the proposed method (\ours).

For \cref{tbl:exp_simple_methods_accuracy}, all methods use same set of retrievals obtained via Contriever with $k{=}5$ on the main query. The prompts employed by each model are detailed in \cref{app:prompting_methods}. To eliminate stochasticity, we set temperature${=}0$ across all models. For \cref{tbl:exp_sota_accuracy}, we evaluate against recently proposed advanced RAG methods: Self-RAG \citep{asai2023self}, RQ-RAG \citep{chan2024rqrag}, and ReAct \citep{yao2023react}. We utilize the official implementations for Self-RAG and RQ-RAG, while adapting ReAct's prompt for the RAG framework (detailed in \cref{app:sota_methods}).

For comparison with state-of-the-art methods, we employ \ours\textsubscript{SubQ}, which retrieves a fresh set of documents for each sub-query, maintaining consistency with the experimental setup of other state-of-the-art approaches.

For both experiments, we employ \textit{accuracy contains} as the evaluation metric. This metric represents a relaxed version of the \textit{exact-match} score, considering a prediction correct when the true answer appears as a substring within the prediction. This approach accounts for variations in answer phrasing across different language models.

\begin{wraptable}{r}{.5\textwidth}
	\centering
	\small
	\caption{\textbf{Information Gain (IG)}: IG at different depths of the reasoning DAG, demonstrating systematic information accumulation across reasoning steps on HotpotQA, validating the structural coherence of the reasoning DAG.}
	\label{tbl:info_gain_depth}
	\begin{tabular}{@{}cccccccc@{}}
		\toprule
		\setlength{\tabcolsep}{2.5pt}
		\textbf{Depth} & 0 & 1 & 2 & 3 & 4 & 5 & 6 \\ 
		\midrule
		\textbf{IG}   & $7.23$ & $8.85$ & $8.91$ & $9.02$ & $9.47$ & $9.75$ & $10.0$ \\ 
		\bottomrule
	\end{tabular}
	\vspace*{-1.2em}
\end{wraptable}

\subsection{Retrieval Effectiveness of \ours}
\label{app:retrieval_precision}
In this experiment, we evaluate the effectiveness of \ours\textsubscript{SubQ}'s reasoning plan nodes, which generate targeted subatomic queries leading to improved retrievals. Using a random sample of $1000$ queries from the HotpotQA \textit{distractor} dataset, which provides golden sentences for evaluation, we assess retrieval performance against ReAct and QD-RAG\textsubscript{SubQ}. As shown in \cref{tbl:precision_recall}, \ours\textsubscript{SubQ} demonstrates superior performance across all threee metrics: Precision, Recall, and Accuracy.

To account for potential fragmentation of golden sentences during embedding generation, we segment the golden sentences into chunks of size $50$. A retrieval is considered successful if any chunk of the golden sentence is present in the retrieval.

\subsection{Information Gain (IG)}
\label{app:information_gain}
We evaluate the effectiveness of \ours's reasoning plan by analyzing the cumulative information gain (IG) across DAG-depth. A valid reasoning plan should demonstrate increasing IG with depth, as each subquery contributes additional relevant information toward answering the main query. Conversely, stagnant or decreasing IG would indicate subqueries at that depth fail to contribute meaningful information. We formally define IG in \cref{eq:information_gain}.

Using the HotpotQA dataset (7,405 queries), we employ GPT-3.5\textsubscript{turbo} to compute and return IG scores at each depth of the reasoning DAG. We showcase the average IG score at each depth in \cref{tbl:info_gain_depth}. The evaluation prompt is
\begin{lstlisting}[frame=single]
Suppose you are an expert in quantifying information gain. Given a main query and a set of subqueries quantify how much knowledge is gained by this whole set of subqueries. Range of information gain: 1-10. 1 being the least and 10 being the most.
The special tag <AI.J> is used to denote the place where the answer to QI.J will be placed.
NOTE: Only output the Information Gain number that can be passed to eval() directly. No explanations or anyother tag. Strictly follow the below template.

Example:

Main Query: What is the DoB of the current President of Finland?
Subqueries: [Q1.1: Who is the current President of Finland?]
Information Gain: 5

Main Query: What is the DoB of the current President of Finland?
Subqueries: [Q1.1: Who is the current President of Finland?, Q2.1: When was <A1.1> born?]
Information Gain: 10

Main Query: What is the DoB of the current President of Finland?
Subqueries: [Q1.1: What is the capital of Finland?]
Information Gain: 0 
\end{lstlisting}

\subsection{Reasoning Plan Generation}
\label{app:reasoning_agent}
While our primary experiments use GPT-4o to generate the reasoning plan, we demonstrate that \ours's performance is not dependent on the large language models. For this, we finetune the Llama3.1-instruct\textsubscript{8b} model with LoRA adapters on a small subset of HotpotQA-\textit{train} dataset (3,700 query-plan pairs), using plans generated by GPT-4 as training data. The results show that a relatively small set of training data is sufficient to finetune a language model to generate structure output required to construct the DAG. 

The finetuning process employs the \textit{Parameter-Efficient Fine-Tuning (PEFT)} library with the following hyperparameters: 5,000 training steps, LoRA rank of 8, LoRA alpha of 16, batch size of 4, and gradient accumulation steps of 2. \cref{tbl:llama3_plan_agent} demonstrates the competitive accuracy of the finetuned model compared to other methods on a subset of HotpotQA-\textit{fullwiki} dataset, confirming that \ours's enhanced performance is not dependent on a large language model.

\end{document}

%% file: fig/teaser.tex
\begin{tikzpicture}
	\begin{axis}[
		ybar,
        clip=false,
		width=\columnwidth,
		height=.73\columnwidth,
		bar width=12pt,
		ylabel={Accuracy (\%)},
		xlabel={},
		ylabel near ticks,
		xtick=data,
		xtick align=inside,
		xtick style={draw=none},
		x tick label style={font=\small, rotate=0, anchor=north},
		ymin=20, ymax=45,
		ytick={20, 25, 30, 35, 40, 45},
		ytick style={draw=none},
		yticklabel style={font=\small},
		nodes near coords,
		nodes near coords style={font=\small, black},
		legend style={at={(0.5,.9)}, anchor=north, legend columns=2, font=\small,draw=none},
		axis x line = bottom,
		axis y line = left,
		axis line style={-},
		enlargelimits=0.3,
		ymajorgrids,
		grid style={dotted, gray},
		symbolic x coords={RAG,Self-RAG,ReAct},
		every node near coord/.append style={font=\scriptsize},
		]

		\addplot [fill=primarycolor,draw=primarycolor,bar shift=-1em,rounded corners=1pt] coordinates {
			(RAG, 25.51)
			(Self-RAG, 34.09)
			(ReAct, 33.15)
		};
		
		\addplot [draw=primarycolor,postaction={pattern=north east lines,pattern color=primarycolor},bar shift=1em,rounded corners=1pt] coordinates {
			(RAG, 31.12)
			(Self-RAG, 37.31)
			(ReAct, 40.44)
		};		
		
        \legend{Vanilla,With Plan$^\ast$}
		
    \draw[|->, thick, black, shift={(2.2em,0)}] 
      (axis cs:RAG,25.51) -- (axis cs:RAG,31.12);

    \draw[|->, thick, black, shift={(2.2em,0)}] 
      (axis cs:Self-RAG,34.09) -- (axis cs:Self-RAG,37.31);

    \draw[|->, thick, black, shift={(2.2em,0)}]
      (axis cs:ReAct,33.18) -- (axis cs:ReAct,40.44);		

    \node[fill=white,text width=8cm,align=center] at (axis cs: Self-RAG,51) {Test-time planning improves RAG};
        
	\end{axis}
\end{tikzpicture}

%% file: fig/reasoning_dag_example.tex
\begin{tikzpicture}[
	node distance=0.8cm and 0.0cm,
	base_style/.style={
		rectangle, 
		rounded corners, 
		minimum width=#1, 
		minimum height=2cm, 
		align=center, 
		draw=black, 
		line width=.75pt,
		fill=highlight,
		text width=#1,
		font=\normalsize
	},
	q_style/.style={ 
		minimum width=#1, 
		minimum height=.4cm, 
		align=left, 
		text width=#1
	},
	arrow_style/.style={->, line width=1.0pt, draw=black!70},
	feature_arrow/.style={->, line width=1.2pt, draw=red!70, dashed}
	]
	
	\node (Q) [q_style=5cm] {};
	\node (dummy) [q_style=11cm] at ($(Q)-(.25,0)$) {\texttt{\textbf{Query:} \\What is the distance between the locations that \\ hosted the last two Men's Cricket World Cup finals?}};
	\node (Q11) [base_style=3.5cm, below left=of Q] {$Q1.1$: Where was the last Men's Cricket \\ World Cup final held?};
	\node (Q12) [base_style=3.5cm, below right=of Q] {$Q1.2$: Where was the second last Men's Cricket \\ World Cup final held?};
	\node (Q21) [base_style=3.5cm, below=of Q11] {$Q2.1$: What are the coordinates of \\ $\langle$A1.1$\rangle$?};
	\node (Q22) [base_style=3.5cm, below=of Q12] {$Q2.2$: What are the coordinates of \\ $\langle$A1.2$\rangle$?};
	\node (Q31) [base_style=4cm, below=1.7cm of $(Q21)!0.5!(Q22)$] {$Q3.1$: What is the distance between $\langle$A2.1$\rangle$ and $\langle$A2.2$\rangle$?};
	
	\draw[arrow_style] (Q11) -- (Q21);
	\draw[arrow_style] (Q12) -- (Q22);
	\draw[arrow_style] (Q21) -- (Q31);
	\draw[arrow_style] (Q22) -- (Q31);
	
	\node[rotate=90] (feat1) at ($(Q21.west)+(-1,0)$) {\textbf{Relevant flow of information}};
	\draw[feature_arrow] ($(Q11.west)+(-.5,0)$) -- ++(0,-5.5cm);
	
	\node[rotate=90] (feat2)  at ($(Q22.east)+(1,0)$) {\textbf{Debuggability \& Backtracking}};
	\draw[feature_arrow,<-] ($(Q12.east)+(.5,0)$) -- ++(0,-5.5cm);
	
	\node(feat3) at ($(Q11)!.5!(Q12)$) {\textbf{Parallelizable subqueries}};
	\draw[feature_arrow,shorten >=2pt] (feat3.west) -- (Q11.east);
	\draw[feature_arrow,shorten >=2pt] (feat3.east) -- (Q12.west);	
	
	\node(feat4) at ($(Q21)!.5!(Q22)$) {\textbf{* Targetted Subquery Retrievals}};
	
\end{tikzpicture}

%% file: fig/tbl_simple_acc.tex
   
\newcommand{\rotate}[1]{\parbox[t]{2mm}{\multirow{5}{*}{\rotatebox[origin=c]{90}{\tiny#1}}}} 
  
\begin{tabular}{llcccc}
	\toprule
	\textbf{~} & \textbf{Method} & \textbf{HotpotQA} & \textbf{StrategyQA} & \textbf{MuSiQue} & \textbf{PopQA} \\
	\midrule
	\rotate{GPT-3.5\textsubscript{turbo}}
	& Vanilla-LLM & $30.45$ & $55.37$ & $\phantom{0}6.53$ & $28.45$ \\
	& Vanilla-RAG & $29.83$ & $54.24$ & $\phantom{0}4.27$ &$32.13$ \\
	& CoT-RAG & $29.96$ & $60.66$ & $\phantom{0}6.50$ &$30.02$ \\
	& QD-RAG & $29.31$ & $59.91$ & $\phantom{0}4.85$ &$36.03$ \\
	& \cellcolor{highlight}\planrag & \cellcolor{highlight}$\mathbf{31.72}$ & \cellcolor{highlight}$\mathbf{62.07}$ & \cellcolor{highlight}$\mathbf{\phantom{0}6.71}$ & \cellcolor{highlight}$\mathbf{38.62}$ \\
	\midrule
	\rotate{Llama2-chat\textsubscript{13B}}
	& Vanilla-LLM & $19.33$ & $44.47$ & $\phantom{0}2.81$ & $22.30$ \\
	& Vanilla-RAG & $22.67$ & $39.11$ & $\phantom{0}2.65$ & $31.25$ \\
	& CoT-RAG & $25.93$ & $29.26$ & $\phantom{0}5.02$ &$34.17$ \\
	& QD-RAG & $26.97$ & $42.79$ & $\phantom{0}4.73$ &$36.88$ \\
	& \cellcolor{highlight}\planrag & \cellcolor{highlight}$\cellcolor{highlight}\mathbf{27.51}$ & \cellcolor{highlight}$\mathbf{54.57}$ & \cellcolor{highlight}$\mathbf{\phantom{0}5.32}$ & \cellcolor{highlight}$\mathbf{41.45}$ \\
	\midrule
	\rotate{Llama3.1-instruct\textsubscript{8B}}
	& Vanilla-LLM & $21.34$ & $48.43$ & $\phantom{0}4.18$ &$23.52$ \\
	& Vanilla-RAG & $25.51$ & $47.36$ & $\phantom{0}4.39$ &$39.1$ \\
	& CoT-RAG & $30.10$ & $59.51$ & $\phantom{0}6.00$ &$37.49$ \\
	& QD-RAG & $28.68$ & $62.38$ & $\phantom{0}2.65$ &$41.13$ \\
    & \cellcolor{highlight}\planrag & \cellcolor{highlight}$\mathbf{31.12}$ & \cellcolor{highlight}$\mathbf{68.16}$ & \cellcolor{highlight}$\mathbf{\phantom{0}6.28}$ & \cellcolor{highlight}$\mathbf{41.51}$ \\
	\bottomrule
\end{tabular}

%% file: fig/token_statistics.tex
	\begin{tikzpicture}
	\begin{axis}[
		width=\figurewidth,
		height=\figureheight,
		scale only axis,	
		ybar=1pt,
		bar width=12pt,
		enlarge x limits=1,
		axis lines*=left,
		axis line style={thick},
		major tick length=0pt,
		minor tick length=0pt,
		ylabel={Log Avg.\ tokens / query},
		symbolic x coords={Input, Output},
		xtick=data,
		ymin=0,
		ymax=2800,        
        %
        ymode=log,
        ytick={50,100,250,500,1000,2000},
        yticklabels={50,100,250,500,1000,2000},
		ymajorgrids=true,
		grid style={dotted, gray},
		xlabel={\strut}
		]
		\addplot[fill=cyan!60,rounded corners=1pt] coordinates {
			(Input, 2194.12)
			(Output, 60.71)
		};
		\addlegendentry{ReAct}
		
		\addplot[fill=purple!60,rounded corners=1pt] coordinates {
			(Input, 2761.74)
			(Output, 74.68)
		};
		\addlegendentry{QD-RAG}
		
		\addplot[fill=green!40,rounded corners=1pt] coordinates {
			(Input, 1063.45)
			(Output, 43.24)
		};
		\addlegendentry{CoT-RAG}
		
		\addplot[postaction={pattern=north east lines,pattern color=primarycolor},rounded corners=1pt] coordinates {
			(Input, 1173.39)
			(Output, 141.36)
		};
		\addlegendentry{\ours}
	\end{axis}
\end{tikzpicture}

%% file: fig/hotpotqa_depth_comparison.tex
\begin{tikzpicture}
	\begin{axis}[
		width=\figurewidth,
		height=\figureheight,
		scale only axis,
		xlabel={Reasoning depth},
		ylabel={Percentage of examples (\%)},
		xmin=-0.2,
		xmax=4.2,
		ymin=0,
		ymax=80,
		xtick={0,1,2,3,4},
		xticklabels={0,1,2,3,$\geq$4},
		ytick={0,20,40,60,80},
		ylabel near ticks,
		grid=major,
		grid style={dotted,gray!30},
		legend style={
			at={(0.98,0.98)},
			anchor=north east,
			cells={anchor=west},
			font=\small
		},
		axis lines*=left,
		axis line style={thick},
		]
		
		\addplot[line width=2pt,cyan!60,smooth,no markers] coordinates {
			(0,1.0)
			(1,9.9)
			(2,46.5)
			(3,18.0)
			(4,24.6)
		};
		\addlegendentry{ReAct}
		
		\addplot[line width=2pt,purple!60,smooth,no markers] coordinates {
			(0,9.0)
			(1,4.6)
			(2,58.5)
			(3,22.7)
			(4,5.2)
		};
		\addlegendentry{QD-RAG}
		

		\addplot[line width=2pt, primarycolor,smooth,no markers,dotted] coordinates {
			(0,0.5)
			(1,12.1)
			(2,75.4)
			(3,10.3)
			(4,0.4)
		};
		\addlegendentry{\ours}
				
	\end{axis}
\end{tikzpicture}

%% file: fig/tbl_sota_acc.tex
\newcommand{\rotate}[2]{\parbox[c]{2mm}{\multirow{#1}{*}{\tikz[baseline]\node[anchor=base,rotate=90,font=\tiny,align=center]{#2};}}}

\begin{tabular}{llcccc}
	\toprule
	\textbf{Base} & \textbf{Method} & \textbf{HotpotQA} & \textbf{StrategyQA} & \textbf{MuSiQue} & \textbf{PopQA} \\
	\midrule
    
    \rotate{3}{Llama2\textsubscript{7B}}
	& RQ-RAG  & $33.78$ & $47.46$ & $13.43$ & $32.66$ \\
	& Self-RAG & $31.48$ & $43.71$ & $7.07$ & $43.67$ \\
	&  \cellcolor{highlight}Plan$^\ast$-Self-RAG & \cellcolor{highlight}$\mathbf{36.65}$ & \cellcolor{highlight}$\mathbf{60.67}$ & \cellcolor{highlight}$\mathbf{14.76}$ & \cellcolor{highlight}$\mathbf{44.06}$ \\
	
	\midrule	
    
    \rotate{2}{Llama2\textsubscript{13B}} 
	& Self-RAG & $34.09$ & $42.75$ & $8.31$ & $45.18$ \\
	& \cellcolor{highlight}Plan$^\ast$-Self-RAG & \cellcolor{highlight}$\mathbf{37.31}$ & \cellcolor{highlight}$\mathbf{62.87}$ & \cellcolor{highlight}$\mathbf{16.72}$  & \cellcolor{highlight}$\mathbf{45.23}$ \\	
	
	\midrule

    \rotate{2}{Llama3.1 \\ instruct\textsubscript{8B}} 
	& ReAct & $33.15$ & $54.67$ & $13.36$ & $35.81$ \\ 
	& \cellcolor{highlight}\ours\textsubscript{SubQ} & \cellcolor{highlight}$\mathbf{40.44}$ & \cellcolor{highlight}$\mathbf{65.45}$ & \cellcolor{highlight}$\mathbf{14.88}$ & \cellcolor{highlight}$\mathbf{41.98}$ \\ 
		
	\bottomrule
\end{tabular}

%% file: arxiv.bbl
\begin{thebibliography}{44}
\providecommand{\natexlab}[1]{#1}
\providecommand{\url}[1]{\texttt{#1}}
\expandafter\ifx\csname urlstyle\endcsname\relax
  \providecommand{\doi}[1]{doi: #1}\else
  \providecommand{\doi}{doi: \begingroup \urlstyle{rm}\Url}\fi

\bibitem[Asai et~al.(2023)Asai, Wu, Wang, Sil, and Hajishirzi]{asai2023self}
Asai, A., Wu, Z., Wang, Y., Sil, A., and Hajishirzi, H.
\newblock {Self-RAG}: {L}earning to retrieve, generate, and critique through
  self-reflection.
\newblock In \emph{The Twelfth International Conference on Learning
  Representations}, 2023.

\bibitem[Borgeaud et~al.(2022)Borgeaud, Mensch, Hoffmann, Cai, Rutherford,
  Millican, Van Den~Driessche, Lespiau, Damoc, Clark, De~Las~Casas, Guy,
  Menick, Ring, Hennigan, Huang, Maggiore, Jones, Cassirer, Brock, Paganini,
  Irving, Vinyals, Osindero, Simonyan, Rae, Elsen, and
  Sifre]{pmlr-v162-borgeaud22a}
Borgeaud, S., Mensch, A., Hoffmann, J., Cai, T., Rutherford, E., Millican, K.,
  Van Den~Driessche, G.~B., Lespiau, J.-B., Damoc, B., Clark, A., De~Las~Casas,
  D., Guy, A., Menick, J., Ring, R., Hennigan, T., Huang, S., Maggiore, L.,
  Jones, C., Cassirer, A., Brock, A., Paganini, M., Irving, G., Vinyals, O.,
  Osindero, S., Simonyan, K., Rae, J., Elsen, E., and Sifre, L.
\newblock Improving language models by retrieving from trillions of tokens.
\newblock In \emph{Proceedings of the 39th International Conference on Machine
  Learning}, volume 162 of \emph{Proceedings of Machine Learning Research},
  pp.\  2206--2240. PMLR, 2022.

\bibitem[Brown et~al.(2020)Brown, Mann, Ryder, Subbiah, Kaplan, Dhariwal,
  Neelakantan, Shyam, Sastry, Askell, et~al.]{brown2020language}
Brown, T., Mann, B., Ryder, N., Subbiah, M., Kaplan, J.~D., Dhariwal, P.,
  Neelakantan, A., Shyam, P., Sastry, G., Askell, A., et~al.
\newblock Language models are few-shot learners.
\newblock In \emph{Advances in Neural Information Processing Systems},
  volume~33, pp.\  1877--1901. Curran Associates, Inc., 2020.

\bibitem[Chan et~al.(2024)Chan, Xu, Yuan, Luo, Xue, Guo, and Fu]{chan2024rqrag}
Chan, C.-M., Xu, C., Yuan, R., Luo, H., Xue, W., Guo, Y., and Fu, J.
\newblock {RQ}-{RAG}: Learning to refine queries for retrieval augmented
  generation.
\newblock In \emph{First Conference on Language Modeling}, 2024.

\bibitem[Chen et~al.(2017)Chen, Fisch, Weston, and Bordes]{chen2017reading}
Chen, D., Fisch, A., Weston, J., and Bordes, A.
\newblock Reading {W}ikipedia to answer open-domain questions.
\newblock In \emph{Proceedings of the 55th Annual Meeting of the Association
  for Computational Linguistics (Volume 1: Long Papers)}, pp.\  1870--1879.
  Association for Computational Linguistics, 2017.

\bibitem[Chen et~al.(2024)Chen, White, Mooney, Payani, Su, and
  Sun]{chen2024tree}
Chen, Z., White, M., Mooney, R., Payani, A., Su, Y., and Sun, H.
\newblock When is tree search useful for llm planning? it depends on the
  discriminator.
\newblock \emph{arXiv preprint arXiv:2402.10890}, 2024.

\bibitem[Drozdov et~al.(2022)Drozdov, Wang, Rahimi, Mccallum, Zamani, and
  Iyyer]{drozdov2022you}
Drozdov, A., Wang, S., Rahimi, R., Mccallum, A., Zamani, H., and Iyyer, M.
\newblock You can’t pick your neighbors, or can you? {W}hen and how to rely
  on retrieval in the {kNN-LM}.
\newblock In \emph{Findings of the Association for Computational Linguistics
  (EMNLP)}, pp.\  2997--3007, 2022.

\bibitem[Geva et~al.(2021)Geva, Khashabi, Segal, Khot, Roth, and
  Berant]{Geva2021DidAU}
Geva, M., Khashabi, D., Segal, E., Khot, T., Roth, D., and Berant, J.
\newblock Did {A}ristotle use a laptop? {A} question answering benchmark with
  implicit reasoning strategies.
\newblock \emph{Transactions of the Association for Computational Linguistics},
  9:\penalty0 346--361, 2021.

\bibitem[Guu et~al.(2020)Guu, Lee, Tung, Pasupat, and Chang]{guu2020retrieval}
Guu, K., Lee, K., Tung, Z., Pasupat, P., and Chang, M.
\newblock Retrieval augmented language model pre-training.
\newblock In \emph{Proceedings of the International Conference on Machine
  Learning}, pp.\  3929--3938. PMLR, 2020.

\bibitem[Hao et~al.(2023)Hao, Gu, Ma, Hong, Wang, Wang, and
  Hu]{hao2023reasoning}
Hao, S., Gu, Y., Ma, H., Hong, J.~J., Wang, Z., Wang, D.~Z., and Hu, Z.
\newblock Reasoning with language model is planning with world model.
\newblock In \emph{Proceedings of the 2023 Conference on Empirical Methods in
  Natural Language Processing}, 2023.

\bibitem[Hsu et~al.(2024)Hsu, Khattab, Finn, and
  Sharma]{hsu2024groundingtryingllmsreinforcement}
Hsu, S., Khattab, O., Finn, C., and Sharma, A.
\newblock Grounding by trying: {LLM}s with reinforcement learning-enhanced
  retrieval.
\newblock \emph{arXiv preprint arxiv:2410.23214}, 2024.

\bibitem[Izacard et~al.(2022)Izacard, Caron, Hosseini, Riedel, Bojanowski,
  Joulin, and Grave]{izacard2022unsupervised}
Izacard, G., Caron, M., Hosseini, L., Riedel, S., Bojanowski, P., Joulin, A.,
  and Grave, E.
\newblock Unsupervised dense information retrieval with contrastive learning.
\newblock \emph{Transactions on Machine Learning Research}, 2022.

\bibitem[Izacard et~al.(2023)Izacard, Lewis, Lomeli, Hosseini, Petroni, Schick,
  Dwivedi-Yu, Joulin, Riedel, and Grave]{izacard2023atlas}
Izacard, G., Lewis, P., Lomeli, M., Hosseini, L., Petroni, F., Schick, T.,
  Dwivedi-Yu, J., Joulin, A., Riedel, S., and Grave, E.
\newblock Atlas: Few-shot learning with retrieval augmented language models.
\newblock \emph{Journal of Machine Learning Research}, 24\penalty0
  (251):\penalty0 1--43, 2023.

\bibitem[Jiang et~al.(2023)Jiang, Xu, Gao, Sun, Liu, Dwivedi-Yu, Yang, Callan,
  and Neubig]{jiang2023active}
Jiang, Z., Xu, F.~F., Gao, L., Sun, Z., Liu, Q., Dwivedi-Yu, J., Yang, Y.,
  Callan, J., and Neubig, G.
\newblock Active retrieval augmented generation.
\newblock In \emph{Proceedings of the 2023 Conference on Empirical Methods in
  Natural Language Processing}, pp.\  7969--7992, 2023.

\bibitem[Leng et~al.(2024)Leng, Portes, Havens, Zaharia, and
  Carbin]{leng2024long}
Leng, Q., Portes, J., Havens, S., Zaharia, M., and Carbin, M.
\newblock Long context rag performance of large language models.
\newblock \emph{arXiv preprint arXiv:2411.03538}, 2024.

\bibitem[Lewis et~al.(2020)Lewis, Perez, Piktus, Petroni, Karpukhin, Goyal,
  K{\"u}ttler, Lewis, Yih, Rockt{\"a}schel, Riedel, and
  Kiela]{lewis2020retrieval}
Lewis, P., Perez, E., Piktus, A., Petroni, F., Karpukhin, V., Goyal, N.,
  K{\"u}ttler, H., Lewis, M., Yih, W.-t., Rockt{\"a}schel, T., Riedel, S., and
  Kiela, D.
\newblock Retrieval-augmented generation for knowledge-intensive {NLP} tasks.
\newblock In \emph{Advances in Neural Information Processing Systems},
  volume~33, pp.\  9459--9474. Curran Associates, Inc., 2020.

\bibitem[Liu et~al.(2024)Liu, Peng, Zhang, Liu, Yin, Cao, and Du]{liu2024ra}
Liu, Y., Peng, X., Zhang, X., Liu, W., Yin, J., Cao, J., and Du, T.
\newblock Ra-isf: Learning to answer and understand from retrieval augmentation
  via iterative self-feedback.
\newblock \emph{arXiv preprint arXiv:2403.06840}, 2024.

\bibitem[Luo et~al.(2023)Luo, Zhang, Chuang, Gong, Kim, Wu, Meng, and
  Glass]{luo2023search}
Luo, H., Zhang, T., Chuang, Y.-S., Gong, Y., Kim, Y., Wu, X., Meng, H.~M., and
  Glass, J.~R.
\newblock Search augmented instruction learning.
\newblock In \emph{Proceedings of the 2023 Conference on Empirical Methods in
  Natural Language Processing}, 2023.

\bibitem[Ma et~al.(2023)Ma, Gong, He, Zhao, and Duan]{ma2023query}
Ma, X., Gong, Y., He, P., Zhao, H., and Duan, N.
\newblock Query rewriting in retrieval-augmented large language models.
\newblock In \emph{Proceedings of the 2023 Conference on Empirical Methods in
  Natural Language Processing}, pp.\  5303--5315, 2023.

\bibitem[Mallen et~al.(2022)Mallen, Asai, Zhong, Das, Hajishirzi, and
  Khashabi]{mallen2023llm_memorization}
Mallen, A., Asai, A., Zhong, V., Das, R., Hajishirzi, H., and Khashabi, D.
\newblock When not to trust language models: {I}nvestigating effectiveness and
  limitations of parametric and non-parametric memories.
\newblock \emph{arXiv preprint arXiv:2212.10511}, 2022.

\bibitem[Mihaylov et~al.(2018)Mihaylov, Clark, Khot, and
  Sabharwal]{mihaylov2018suit}
Mihaylov, T., Clark, P., Khot, T., and Sabharwal, A.
\newblock Can a suit of armor conduct electricity? {A} new dataset for open
  book question answering.
\newblock \emph{arXiv preprint arXiv:1809.02789}, 2018.

\bibitem[Pal et~al.(2023)Pal, Umapathi, and Sankarasubbu]{pal2023med}
Pal, A., Umapathi, L.~K., and Sankarasubbu, M.
\newblock Med-halt: Medical domain hallucination test for large language
  models.
\newblock In \emph{Proceedings of the 27th Conference on Computational Natural
  Language Learning (CoNLL)}, pp.\  314--334, 2023.

\bibitem[Patel et~al.(2022)Patel, Mishra, Parmar, and Baral]{patel2022question}
Patel, P., Mishra, S., Parmar, M., and Baral, C.
\newblock Is a question decomposition unit all we need?
\newblock In \emph{Proceedings of the 2022 Conference on Empirical Methods in
  Natural Language Processing}, pp.\  4553--4569, 2022.

\bibitem[Petroni et~al.(2020)Petroni, Lewis, Piktus, Rockt{\"a}schel, Wu,
  Miller, and Riedel]{petroni2020how}
Petroni, F., Lewis, P., Piktus, A., Rockt{\"a}schel, T., Wu, Y., Miller, A.~H.,
  and Riedel, S.
\newblock How context affects language models' factual predictions.
\newblock In \emph{Automated Knowledge Base Construction}, 2020.

\bibitem[Ram et~al.(2023)Ram, Levine, Dalmedigos, Muhlgay, Shashua,
  Leyton-Brown, and Shoham]{ram2023context}
Ram, O., Levine, Y., Dalmedigos, I., Muhlgay, D., Shashua, A., Leyton-Brown,
  K., and Shoham, Y.
\newblock In-context retrieval-augmented language models.
\newblock \emph{Transactions of the Association for Computational Linguistics},
  11:\penalty0 1316--1331, 2023.

\bibitem[Ranaldi et~al.(2024)Ranaldi, Valentino, and
  Freitas]{ranaldi2024eliciting}
Ranaldi, L., Valentino, M., and Freitas, A.
\newblock Eliciting critical reasoning in retrieval-augmented language models
  via contrastive explanations.
\newblock \emph{arXiv preprint arXiv:2410.22874}, 2024.

\bibitem[Schick et~al.(2024)Schick, Dwivedi-Yu, Dess{\`\i}, Raileanu, Lomeli,
  Hambro, Zettlemoyer, Cancedda, and Scialom]{schick2024toolformer}
Schick, T., Dwivedi-Yu, J., Dess{\`\i}, R., Raileanu, R., Lomeli, M., Hambro,
  E., Zettlemoyer, L., Cancedda, N., and Scialom, T.
\newblock Toolformer: {L}anguage models can teach themselves to use tools.
\newblock In \emph{Advances in Neural Information Processing Systems},
  volume~36. Curran Associates, Inc., 2024.

\bibitem[Shuster et~al.(2021)Shuster, Poff, Chen, Kiela, and
  Weston]{shuster2021retrieval}
Shuster, K., Poff, S., Chen, M., Kiela, D., and Weston, J.
\newblock Retrieval augmentation reduces hallucination in conversation.
\newblock \emph{arXiv preprint arXiv:2104.07567}, 2021.

\bibitem[Sun et~al.(2024)Sun, Xu, Tang, Wang, Lin, Gong, Ni, Shum, and
  Guo]{sun2024thinkongraph}
Sun, J., Xu, C., Tang, L., Wang, S., Lin, C., Gong, Y., Ni, L., Shum, H.-Y.,
  and Guo, J.
\newblock Think-on-graph: Deep and responsible reasoning of large language
  model on knowledge graph.
\newblock In \emph{The Twelfth International Conference on Learning
  Representations}, 2024.

\bibitem[Tang \& Yang(2024)Tang and Yang]{tang2024multihoprag}
Tang, Y. and Yang, Y.
\newblock Multihop-{RAG}: Benchmarking retrieval-augmented generation for
  multi-hop queries.
\newblock In \emph{First Conference on Language Modeling}, 2024.

\bibitem[Torfi et~al.(2020)Torfi, Shirvani, Keneshloo, Tavaf, and
  Fox]{torfi2020natural}
Torfi, A., Shirvani, R.~A., Keneshloo, Y., Tavaf, N., and Fox, E.~A.
\newblock Natural language processing advancements by deep learning: A survey.
\newblock \emph{arXiv preprint arXiv:2003.01200}, 2020.

\bibitem[Trivedi et~al.(2022)Trivedi, Balasubramanian, Khot, and
  Sabharwal]{trivedi2022musique}
Trivedi, H., Balasubramanian, N., Khot, T., and Sabharwal, A.
\newblock Musique: {M}ultihop questions via single-hop question composition.
\newblock \emph{Transactions of the Association for Computational Linguistics},
  10:\penalty0 539--554, 2022.

\bibitem[Wang et~al.(2024{\natexlab{a}})Wang, Ping, Mcafee, Xu, Li, Shoeybi,
  and Catanzaro]{pmlr-v235-wang24bd}
Wang, B., Ping, W., Mcafee, L., Xu, P., Li, B., Shoeybi, M., and Catanzaro, B.
\newblock {I}nstruct{R}etro: Instruction tuning post retrieval-augmented
  pretraining.
\newblock In \emph{Proceedings of the 41st International Conference on Machine
  Learning}, volume 235 of \emph{Proceedings of Machine Learning Research},
  pp.\  51255--51272. PMLR, 2024{\natexlab{a}}.

\bibitem[Wang et~al.(2023)Wang, Wei, Schuurmans, Le, Chi, Narang, Chowdhery,
  and Zhou]{wang2023selfconsistency}
Wang, X., Wei, J., Schuurmans, D., Le, Q.~V., Chi, E.~H., Narang, S.,
  Chowdhery, A., and Zhou, D.
\newblock Self-consistency improves chain of thought reasoning in language
  models.
\newblock In \emph{The Eleventh International Conference on Learning
  Representations}, 2023.

\bibitem[Wang et~al.(2024{\natexlab{b}})Wang, Liu, Lin, Li, Ma, and
  Liang]{wang2024rat}
Wang, Z., Liu, A., Lin, H., Li, J., Ma, X., and Liang, Y.
\newblock {RAT}: {R}etrieval augmented thoughts elicit context-aware reasoning
  in long-horizon generation.
\newblock \emph{arXiv preprint arXiv:2403.05313}, 2024{\natexlab{b}}.

\bibitem[Wei et~al.(2022)Wei, Wang, Schuurmans, Bosma, Xia, Chi, Le, Zhou,
  et~al.]{wei2022chain}
Wei, J., Wang, X., Schuurmans, D., Bosma, M., Xia, F., Chi, E., Le, Q.~V.,
  Zhou, D., et~al.
\newblock Chain-of-thought prompting elicits reasoning in large language
  models.
\newblock In \emph{Advances in Neural Information Processing Systems},
  volume~35, pp.\  24824--24837. Curran Associates, Inc., 2022.

\bibitem[Welleck et~al.(2024)Welleck, Bertsch, Finlayson, Schoelkopf, Xie,
  Neubig, Kulikov, and Harchaoui]{welleck2024decoding}
Welleck, S., Bertsch, A., Finlayson, M., Schoelkopf, H., Xie, A., Neubig, G.,
  Kulikov, I., and Harchaoui, Z.
\newblock From decoding to meta-generation: Inference-time algorithms for large
  language models.
\newblock \emph{arXiv preprint arXiv:2406.16838}, 2024.

\bibitem[Yang et~al.(2018)Yang, Qi, Zhang, Bengio, Cohen, Salakhutdinov, and
  Manning]{yang2018hotpotqa}
Yang, Z., Qi, P., Zhang, S., Bengio, Y., Cohen, W.~W., Salakhutdinov, R., and
  Manning, C.~D.
\newblock {HotpotQA}: A dataset for diverse, explainable multi-hop question
  answering.
\newblock In \emph{Proceedings of the 2018 Conference on Empirical Methods in
  Natural Language Processing}, 2018.

\bibitem[Yao et~al.(2023)Yao, Zhao, Yu, Du, Shafran, Narasimhan, and
  Cao]{yao2023react}
Yao, S., Zhao, J., Yu, D., Du, N., Shafran, I., Narasimhan, K., and Cao, Y.
\newblock Re{A}ct: {S}ynergizing reasoning and acting in language models.
\newblock In \emph{International Conference on Learning Representations}, 2023.

\bibitem[Yao et~al.(2024)Yao, Yu, Zhao, Shafran, Griffiths, Cao, and
  Narasimhan]{yao2024tree}
Yao, S., Yu, D., Zhao, J., Shafran, I., Griffiths, T., Cao, Y., and Narasimhan,
  K.
\newblock Tree of thoughts: Deliberate problem solving with large language
  models.
\newblock In \emph{Advances in Neural Information Processing Systems},
  volume~36. Curran Associates, Inc., 2024.

\bibitem[Yogatama et~al.(2021)Yogatama, de~Masson~d’Autume, and
  Kong]{yogatama2021adaptive}
Yogatama, D., de~Masson~d’Autume, C., and Kong, L.
\newblock Adaptive semiparametric language models.
\newblock \emph{Transactions of the Association for Computational Linguistics},
  9:\penalty0 362--373, 2021.

\bibitem[Zhao et~al.(2024)Zhao, Liu, Wu, Li, Yang, Shu, Xu, Dai, Zhao, Mai,
  et~al.]{zhao2024revolutionizing}
Zhao, H., Liu, Z., Wu, Z., Li, Y., Yang, T., Shu, P., Xu, S., Dai, H., Zhao,
  L., Mai, G., et~al.
\newblock Revolutionizing finance with llms: An overview of applications and
  insights.
\newblock \emph{arXiv preprint arXiv:2401.11641}, 2024.

\bibitem[Zhao et~al.(2023)Zhao, Zhou, Li, Tang, Wang, Hou, Min, Zhang, Zhang,
  Dong, et~al.]{zhao2023survey}
Zhao, W.~X., Zhou, K., Li, J., Tang, T., Wang, X., Hou, Y., Min, Y., Zhang, B.,
  Zhang, J., Dong, Z., et~al.
\newblock A survey of large language models.
\newblock \emph{arXiv preprint arXiv:2303.18223}, 2023.

\bibitem[Zhou et~al.(2023)Zhou, Sch{\"a}rli, Hou, Wei, Scales, Wang,
  Schuurmans, Cui, Bousquet, Le, and Chi]{zhou2023leasttomost}
Zhou, D., Sch{\"a}rli, N., Hou, L., Wei, J., Scales, N., Wang, X., Schuurmans,
  D., Cui, C., Bousquet, O., Le, Q.~V., and Chi, E.~H.
\newblock Least-to-most prompting enables complex reasoning in large language
  models.
\newblock In \emph{The Eleventh International Conference on Learning
  Representations}, 2023.

\end{thebibliography}
